\documentclass[11pt]{article}
\usepackage{acl}
\usepackage{svg}
\usepackage{times}

\usepackage[T5]{fontenc}
\usepackage[utf8]{inputenc}
\usepackage{graphicx}
\usepackage{subcaption}
\usepackage{float}

\usepackage{enumitem} 

\usepackage{multirow}
\usepackage{booktabs} 

\usepackage{longtable}
\usepackage{xcolor}
\usepackage{colortbl} 
\definecolor{greenhead}{HTML}{D9EAD3}
\definecolor{yellowhead}{HTML}{FFF2CC}
\definecolor{redhead}{HTML}{FCE5CD}

\title{ViGoEmotions: A Benchmark Dataset For Fine-grained Emotion Detection on Vietnamese Texts}

\author{Hung Quang Tran$^{1,2}$,
    Nam Tien Pham$^{1,2}$,
    Son T. Luu$^{1,2}$, 
    Kiet Van Nguyen$^{1,2,}$\Thanks{Corresponding author: kietnv@uit.edu.vn}\\
    $^{1}$Faculty of Information Science and Engineering, University of Information Technology, \\Ho Chi Minh City, Vietnam\\
    $^{2}$Vietnam National University, Ho Chi Minh City, Vietnam\\
  \texttt{\{22540008,22540012\}@gm.uit.edu.vn, \{sonlt,kietnv\}@uit.edu.vn} \\
  }

\begin{document}
\maketitle
\begin{abstract}
Emotion classification plays a significant role in emotion prediction and harmful content detection. Recent advancements in NLP, particularly through large language models (LLMs), have greatly improved outcomes in this field. This study introduces ViGoEmotions - a Vietnamese emotion corpus comprising 20,664 social media comments in which each comment is classified into 27 fine-grained distinct emotions. To evaluate the quality of the dataset and its impact on emotion classification, eight pre-trained Transformer-based models were evaluated under three preprocessing strategies: preserving original emojis with rule-based normalization, converting emojis into textual descriptions, and applying ViSoLex, a model-based lexical normalization system. Results show that converting emojis into text often improves the performance of several BERT-based baselines, while preserving emojis yields the best results for ViSoBERT and CafeBERT. In contrast, removing emojis generally leads to lower performance. ViSoBERT achieved the highest Macro F1-score of 61.50\% and Weighted F1-score of 63.26\%. Strong performance was also observed from CafeBERT and PhoBERT. These findings highlight that while the proposed corpus can support diverse architectures effectively, preprocessing strategies and annotation quality remain key factors influencing downstream performance.

\end{abstract}

\section{Introduction}
\label{intro}
Understanding human emotions is inherently complex, as even brief comments can express multiple emotional states simultaneously. Accurate emotion recognition in text is essential for tasks such as sentiment analysis, emotion prediction, and harmful content detection, requiring approaches that go beyond single-label classification.

Most Vietnamese corpora adopt basic emotion models such as Ekman's six universal emotions \cite{ekman_are_1992}. More comprehensive frameworks like Plutchik’s wheel \cite{plutchik_nature_1988}, Russell’s circumplex model \cite{russell_circumplex_1980}, and Barrett’s dimensional theory \cite{barrett_valence_2006} highlight emotion complexity via valence-arousal axes or hierarchical relations. However, recent advances in psychology demonstrate that emotions can be modeled in a complex semantic space, such that they respond to a diverse array of representations via computational techniques \cite{cowen2019mapping}. These theories support us in adopting the 27-emotion schema inspired by GoEmotions \cite{demszky_goemotions_2020}.

\begin{table*}[ht!]
    \centering 
    \resizebox{\textwidth}{!}{
    {\small
    \begin{tabular}{|c|p{5cm}|p{5cm}|>{\raggedright\arraybackslash}p{3cm}|}
    
    \hline
    
    \textbf{No.} & \textbf{Vietnamese sentences} & \textbf{English translation} & \textbf{Emotion} \\ \hline
    1 & cần tìm gấp bạn trai đáng yêu như anh này trời đất ơi =)) & I urgently need to find a cute boyfriend like this guy =)) & amusement, desire \\ \hline
    2 & hóng mãi cuối cùng bị cho leo cây \includesvg[scale=0.015]{emojis/unamused-face} & Been waiting forever, only to be stood up in the end \includesvg[scale=0.015]{emojis/unamused-face} & disappointment, annoyance \\  \hline
    3 & tội cả con quá \includesvg[scale=0.015]{emojis/loudly-crying-face}\includesvg[scale=0.015]{emojis/loudly-crying-face} một lũ vô nhân tính đã hủy hoại những mầm non tương lai . & Poor child \includesvg[scale=0.015]{emojis/loudly-crying-face}\includesvg[scale=0.015]{emojis/loudly-crying-face} A group of heartless people has destroyed the future seedlings. & sadness, grief, anger \\ \hline
    \end{tabular}
    } 
    }
    \caption{Translation and emotion annotation of Vietnamese sentences. More examples can be found in Table \ref{tab:example_more} in the Appendix \ref{appendix_sample}.} 
    \label{tab:example}
\end{table*}

Recently, emotion recognition in Vietnamese remains underdeveloped due to the limited scale and granularity of existing datasets, which often rely on basic emotion taxonomies and lack diversity in real-world expressions. Moreover, few studies have leveraged LLMs for Vietnamese emotion annotation. Additionally, constructing a high-quality dataset requires a significant investment of time and resources. To reduce the cost of annotation but preserve the quality, we leverage the large language models (LLMs) like Gemini-flash, a lightweight variant from the Gemini Flash series \cite{google2023gemini}, Llama 3 \cite{touvron2023llama}, Gemma \cite{gemmateam2025gemma3technicalreport} to assist the annotation process since LLMs demonstrate strong ability in natural language understanding, enabling more nuanced emotion classification.
\textcolor{black} {These models represent different architectural families, reducing bias from relying on a single LLM and enabling cross-model comparison. Gemini-flash is optimized for speed and lightweight deployment, while Llama 3 and Gemma are open-source, making them practical for large-scale annotation. }
Although LLMs show potential in data annotation by optimizing time and cost, they still need human verification to ensure accuracy and consistency, especially for complex tasks like fine-grained label emotion detection. 

To address these gaps, we introduce ViGoEmotions - a comprehensive Vietnamese emotion dataset of 20,664 social media comments as \textcolor{black}{shown in} Table \ref{tab:example}, combining the UIT-VSMEC corpus \cite{ho_emotion_2020} (6,921 unique entries) with 13,743 newly collected comments from platforms such as Facebook, YouTube, Reddit, TikTok, Threads, and X (formerly Twitter). ViGoEmotions is annotated by a LLM-Human annotation procedure, in which each comment is annotated with one or more of 27 emotion categories or marked as neutral, using three popular \textcolor{black}{LLMs: Gemini 2.0 Flash, Meta-Llama-3-70B, and Gemma 3}, and revised by humans to determine the final labels. Then, we conduct experiments using eight Vietnamese and multilingual transformer-based language models: mBERT \cite{devlin-etal-2019-bert}, XLM-R \cite{conneau2019unsupervised}, PhoBERT-base-v2 \cite{nguyen_phobert_2020}, ViBERT \cite{bui2020improving}, CafeBERT \cite{do-etal-2024-vlue}, ViSoBERT \cite{nguyen_visobert_2023}, BARTpho \cite{tran2022bartpho}, and ViT5 \cite{phan2022vit5} to evaluate the quality and applicability of the dataset in practice. To sum up, this research makes the following key contributions:

\begin{itemize}
    \item \textbf{Dataset Creation}: We construct a large-scale Vietnamese emotion dataset annotated with 27 detailed emotion categories.
    \item \textbf{LLM Annotation}: We apply cutting-edge LLMs for emotion annotation, demonstrating their effectiveness in capturing nuanced emotional expressions.
    \item \textbf{Benchmark Evaluation}: We evaluated the dataset using Transformer-based model architectures to demonstrate its generalizability and utility in emotion recognition tasks.
\end{itemize}
\section{Related Work}
\label{sec:relatedwork}
\subsection{Emotion Datasets}
Emotion classification in Vietnamese NLP has long suffered from limited resources. Existing datasets tend to focus on sentiment polarity or basic emotions, lacking the depth needed for nuanced applications. For example, 
UIT-VSFC \cite{nguyen_uit-vsfc_2018} provides over 10,000 student feedback entries labeled with positive, neutral, or negative sentiment. UIT-ViSFD \cite{luc_phan_sa2sl_2021} and UIT-ViOCD \cite{nguyen_vietnamese_2021} focus on product reviews and complaints, respectively, but remain limited in emotional scope. On the other hand, UIT-VSMEC \cite{ho_emotion_2020} offers 6,927 social media comments labeled with six basic emotions and one "other" category, making it the most relevant resource for Vietnamese emotion classification to date. However, it remains small and coarse-grained.

GoEmotions \cite{demszky_goemotions_2020}, with 58,000 English Reddit comments and 27 emotion classes, demonstrates the value of fine-grained labeling. It serves as a structural inspiration for our extended corpus. More broadly, recent multilingual efforts such as BRIGHTER \cite{muhammad_brighter_2025} have introduced large-scale human-annotated emotion datasets across 28 languages, focusing on six basic emotions with intensity scores. While BRIGHTER emphasizes low-resource languages and includes some Asian languages, it remains limited to coarse-grained emotion labels. Our work extends this direction by targeting Vietnamese social media and adopting a fine-grained 27-emotion taxonomy.

\subsection{LLM-based Annotation}
LLMs have shown strong potential in affective computing. Welivita et al. \cite{welivita_are_2024} found that GPT-4, Gemini, Llama 2, and Mixtral could match or exceed human performance in classifying 32 fine-grained emotions. LLaMA-Emotion \cite{niu_text_2024} further improved performance by incorporating multimodal inputs and instruction tuning. Besides, Buscemi et al. \cite{buscemi_chatgpt_2024} demonstrated the multilingual capabilities of LLMs in emotion classification across ten languages, including Vietnamese. EmoLLMs \cite{liu_emollms_2024}, trained on 234k emotion-rich samples, achieved state-of-the-art results on AEB benchmarks. 

Building on this, we leverage Gemini Flash, Gemma, and Llama 3 to annotate Vietnamese social media data into 27 emotion categories, demonstrating the feasibility of LLM-driven corpus construction in low-resource settings.
\section{Dataset}
\label{sec:dataset}

\begin{figure*}[ht]
    \centering
    \includegraphics[width=\linewidth]{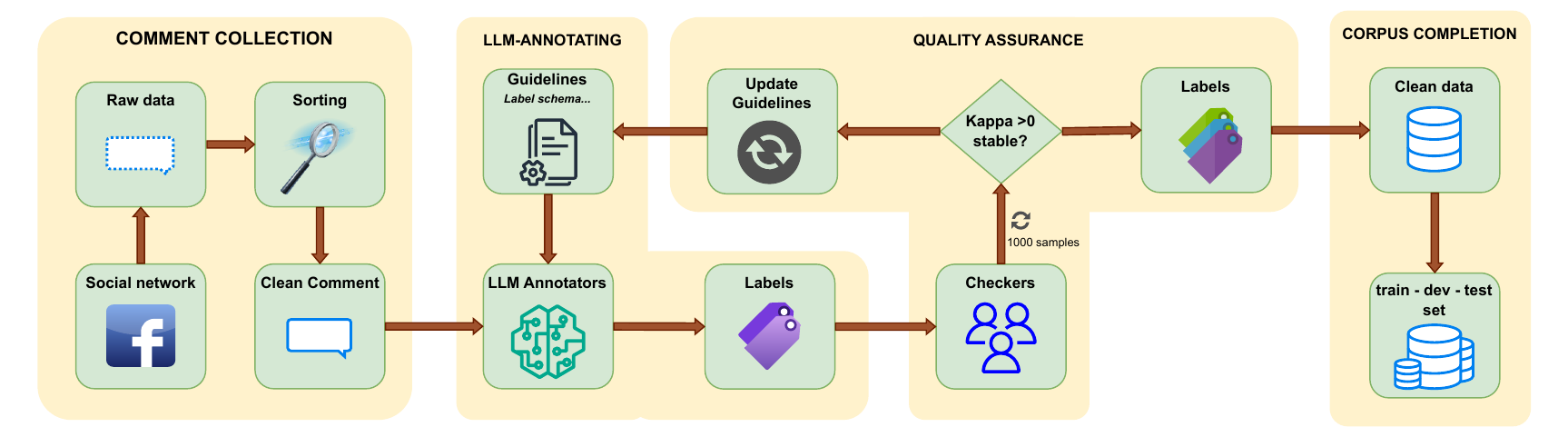}
    \caption{Corpus building process.}
    \label{fig:corpus-building}
\end{figure*}

\begin{figure*}[ht]
	\centering
	\includegraphics[width=0.8\linewidth]{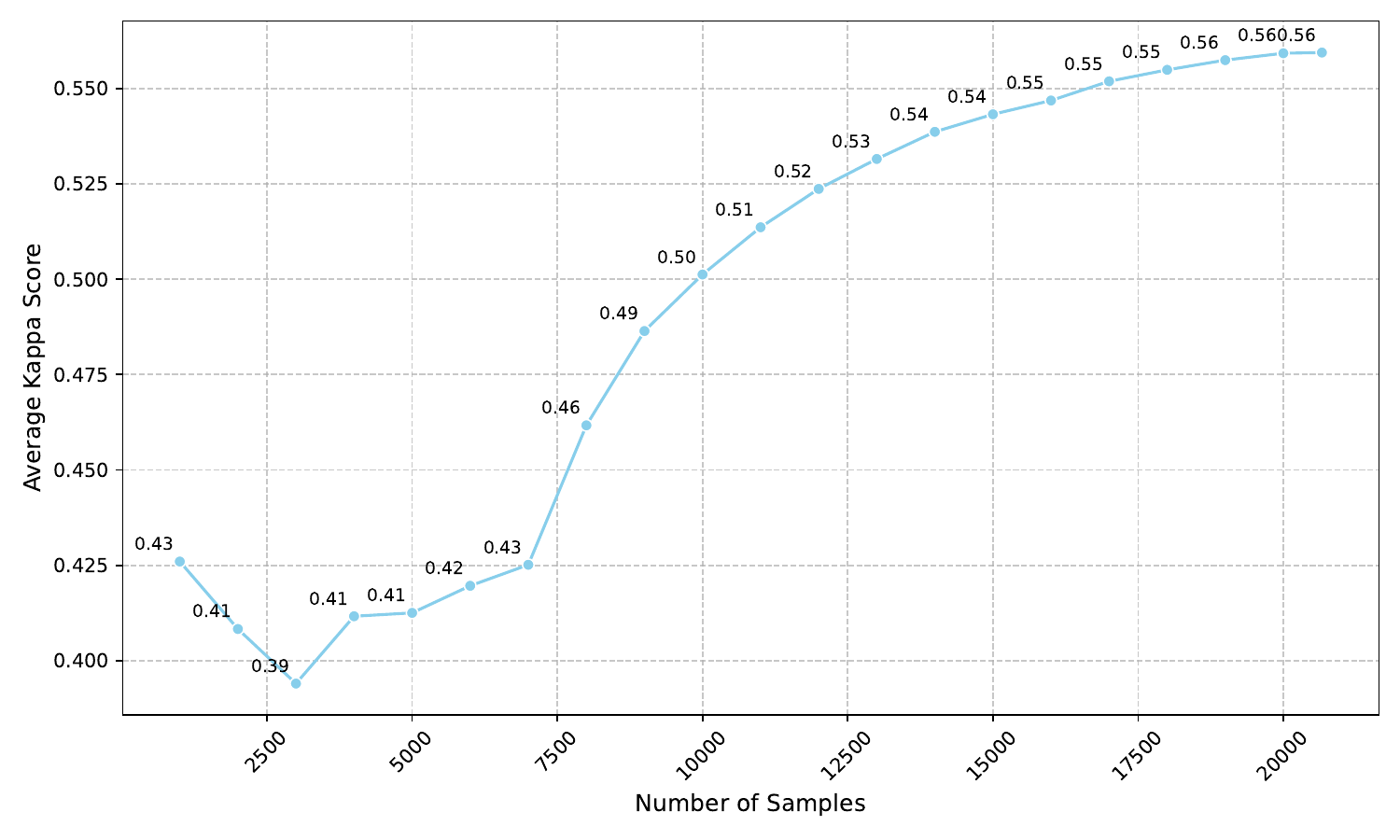}
	\caption{Average Cohen's Kappa Score Over Data Size}
	\label{fig:average-kappa-over-datasize}
\end{figure*}

Figure \ref{fig:corpus-building} illustrates the overall dataset construction pipeline. The process starts with crawling comments from social media to gather raw data, followed by pre-processing to remove noise, duplicates, and irrelevant content. Cleaned data is then annotated by large language models (LLMs) according to predefined annotation guidelines. Annotated data is manually reviewed by human annotators for consistency. Inter-annotator agreement is periodically checked. \textcolor{black}{In each verification round, we compute the average Cohen’s Kappa across all 28 categories. These results will be presented in Figure} \ref{fig:average-kappa-over-datasize}. The annotation guidelines are refined when necessary, and verified data batches are gradually added to the final corpus.

\subsection{Data Collection}
The ViGoEmotions corpus was constructed from two main sources: an existing dataset and newly collected data from various social media platforms. The details are described as follows: 

\begin{itemize}
\item \textbf{Newly collected social media comments (Facebook, Reddit, Thread, TikTok, X, YouTube)}: A total of 53,890 comments were crawled from 144 public posts across diverse domains such as daily life, news, entertainment, and personal narratives. After filtering out duplicated, off-topic, or irrelevant content (e.g., containing only tags, links, or ads), 13,743 high-quality comments were retained for annotation.

\item \textbf{UIT-VSMEC dataset} \cite{ho_emotion_2020}: In addition to new data, we reused the raw comments from the UIT-VSMEC dataset. Although this dataset was pre-annotated with six emotion categories, we re-annotated its comments following our refined taxonomy of 27 fine-grained emotion categories.
\end{itemize}

\subsection{Taxonomy of Emotions and Annotation Guidelines}

Our taxonomy builds upon the GoEmotions label set \cite{demszky_goemotions_2020}, adapted to better suit the objectives of this study. The annotation framework classifies emotions into 27 categories and one neutral label, organized into positive, ambiguous, and negative groups, while considering intensity and contextual dependency. The annotation guidelines were iteratively refined through several annotation trials. In a pilot with 140 samples labeled by two annotators, label distributions and inter-annotator agreement \cite{cohen1960coefficient} were analyzed to identify ambiguities. Based on these results, guidelines were adjusted to clarify label definitions, emoji interpretations, and combinations of commonly co-occurring emotions. More details about the taxonomy of emotion are demonstrated in Appendix \ref{appendix_annotation_guideline}.

Subsequently, these updated guidelines were applied to LLM annotation trials, with outputs evaluated against the same metrics and refined further. This iterative refinement ensures that the annotation framework remains consistent and effective across both human and machine annotations. Three LLMs — \textcolor{black}{Gemini 2.0 Flash, Meta-Llama-3-70B, and Gemma 3}— were employed for automated annotation, guided by structured prompts designed from our guidelines. \textcolor{black}{The models were run with a unified configuration: temperature = \texttt{0.5}, top\_p = \texttt{0.9}, and max\_tokens sampled uniformly within the range \texttt{45–50}.}
Their outputs were not directly used as final labels; instead, three independent human annotators reviewed the candidate labels proposed by LLMs. \textcolor{black}{The annotators were undergraduate students and native Vietnamese speakers; other personal details were not gathered to protect privacy.} Each annotator removed irrelevant suggestions and added missing ones if necessary. The final reference labels were obtained from the intersection of annotators’ decisions, ensuring that the corpus reflects human consensus. \textcolor{black}{In cases where a label did not appear in the intersection of all annotators’ decisions, the final choice was determined by majority agreement. If all three annotators proposed completely different labels, the sample would be reviewed and the guidelines updated accordingly.} The details about the LLMs' prompts are shown in Appendix \ref{appendix_llm_prompt}.

\subsection{Corpus Evaluation}

\begin{figure}[h!]
    \centering
    \includegraphics[width=\linewidth]{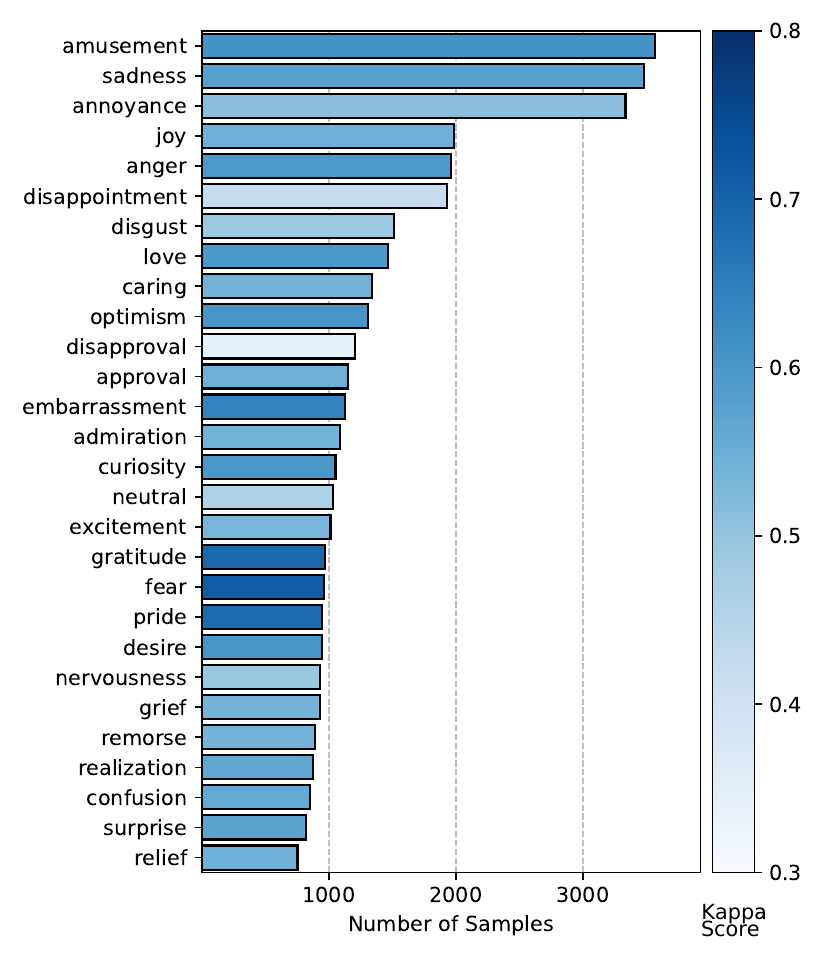}
    \caption{Color indicates Kappa Score across emotion categories. Bars are sorted in descending order by number of samples.}
    \label{fig:quantity_n_kappa}
\end{figure}

To assess the agreement among annotators, we employed the Cohen’s Kappa coefficient (\(\kappa\)) consensus measure, which has been utilized in other notable works, such as UIT-VSMEC and GoEmotions. In our setup with three annotators, the final (\(\kappa\)) score for each category is computed as the average of the pairwise agreements, i.e., $\kappa = \frac{\kappa_{12}+\kappa_{23}+\kappa_{13}}{3}$. As shown in Figure \ref{fig:quantity_n_kappa}, the color scale reflects the Kappa Score, which measures inter-annotator agreement. The bars are sorted by the number of final samples per emotion category, which were retained after applying inter-annotator agreement filtering. Notably, categories such as \textit{fear}, \textit{gratitude}, and \textit{pride} exhibit higher agreement (darker bars), while \textit{disapproval} shows the lowest agreement despite a moderate sample size. Emotion categories such as \textit{amusement} and \textit{sadness} have the largest number of samples, whereas categories like \textit{relief} and \textit{surprise}  are relatively underrepresented. This sorting emphasizes the relationship between agreement levels and final dataset size. For a complete overview of individual Kappa Scores, please refer to Table \ref{tab:kappa_scores} in the Appendix \ref{appendix_kappa}.


\textcolor{black}{To further examine how much human annotators adjusted the labels originally proposed by LLMs, we compare LLM-only labels with human-verified labels. Based on the comparison presented in Table} \ref{tab:llm_human_comparison}, \textcolor{black}{the LLM-generated labels match the human-verified labels in only 40.82\% of the cases. The remaining instances involve partial or complete disagreement, including cases where annotators needed to remove irrelevant labels (16.74\%), add missing labels (22.48\%), or substantially revise the label set (19.96\%). These results show that LLMs reach only partial agreement with human annotators. LLM outputs are not fully aligned with human judgment—especially in socially ambiguous tasks. Therefore, human verification remains necessary to ensure reliable and accurate labeling.}

\begin{table*}[ht]
	\centering
	\caption{Comparison between LLM-assigned labels and human-verified labels.}
	\begin{tabular}{lrr}
		\hline
		\textbf{Category} & \textbf{Count} & \textbf{Percentage} \\
		\hline
		Exact match & 8436 & 40.82\% \\
		Updated (added and removed labels) & 4124 & 19.96\% \\
		Only removed (human $<$ LLM) & 3459 & 16.74\% \\
		Only added (human $>$ LLM) & 4645 & 22.48\% \\
		\hline
	\end{tabular}
	\label{tab:llm_human_comparison}
\end{table*}


\subsection{Corpus Analysis}

\begin{figure*}[ht!]
  \centering
  \includegraphics[width=.9\textwidth]{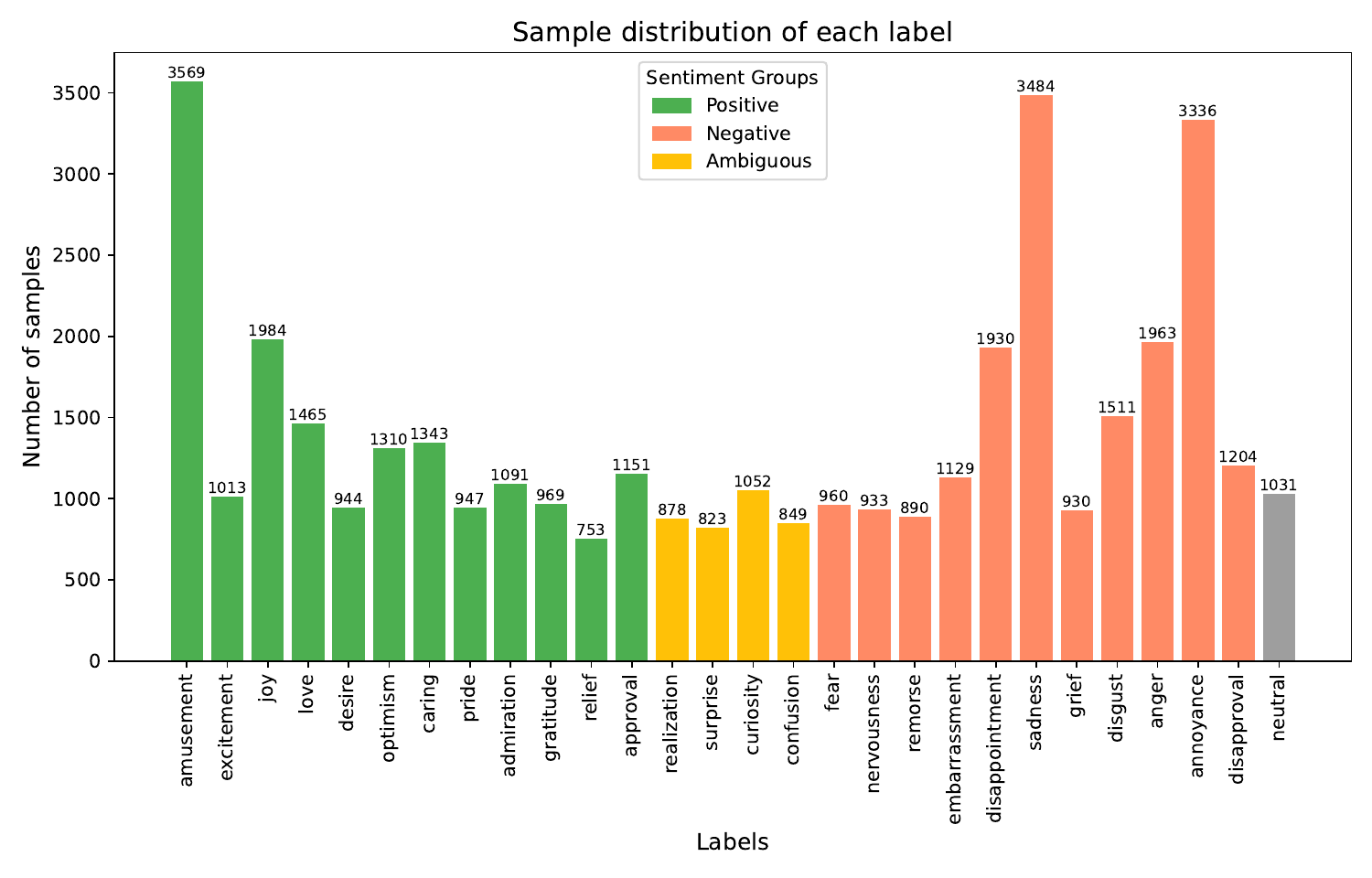}
  \caption{Labels Distribution in the ViGoEmotions}
  \label{fig:LabelsDistribution}
\end{figure*}

\textbf{Emotional Correlation}: Figure \ref{fig:LabelsDistribution} illustrates the label distribution in the multilabel dataset, grouped into four sentiment categories (Positive, Negative, Ambiguous, and Neutral). The bar heights represent the frequency of each label, while the colors highlight their sentiment grouping.
To better understand the semantic relationships among emotion categories, the Pearson correlation coefficient was computed between all emotion pairs based on their co-occurrence patterns, which are illustrated in Figure \ref{fig:pearson_correlation} in the Appendix \ref{appendix_correation}. This analysis helps reveal latent emotional grouping and contextual overlaps - an essential factor when dealing with noisy, multi-emotion social media content. For example, joy is most associated with excitement, love, and amusement, forming a positive emotion cluster. On the other hand, anger, annoyance, and disgust form a tight negative cluster. Additionally, emotions such as curiosity and confusion span both effective poles, frequently co-occurring with both positive and negative categories, highlighting their bridging nature in discourse (to investigate the joint overlap between emotions, we extract the top-5 most correlated emotions per emotion label, summarized in Table \ref{tab:emotion_correlation} from the Appendix \ref{appendix_correation}).

\textbf{Emotion-Specific Word Distribution}: We extracted and visualized the top 5 most representative words for each of the emotion labels in the training set using a frequency-based word cloud approach for better understanding the lexical patterns associated with each emotion category. The results are summarized in Table \ref{tab:top5-words} from the Appendix \ref{appendix_topword}. 

\section{Baseline Models}
\label{sec:method}

To evaluate the challenge of the constructed dataset, we utilized eight Vietnamese and multilingual transformer-based language models.
These models were selected for their effectiveness in Vietnamese NLP and diverse pre-training corpora, which contribute to complementary performance across various domains.

\begin{itemize}
    \item \textbf{mBERT} \cite{devlin-etal-2019-bert}: Multilingual BERT trained on Wikipedia in 104 languages. Though not Vietnamese-specific, it offers reasonable performance through cross-lingual transfer.
    
    \item \textbf{XLM-R} \cite{conneau2019unsupervised}: A robust multilingual model trained on CommonCrawl data in 100 languages, effective for both cross-lingual and low-resource settings.
    
    \item \textbf{PhoBERT} \cite{nguyen_phobert_2020}: Based on RoBERTa and trained on 140GB of Vietnamese text, combining Wikipedia, news, and OSCAR-2301 corpora, strong in formal language processing.
    
    \item \textbf{ViBERT} \cite{bui2020improving}: BERT-based model pre-trained on general Vietnamese corpora, providing solid performance across various tasks.
    
    \item \textbf{BARTpho} \cite{tran2022bartpho}: A BART-based model fine-tuned for Vietnamese generation and understanding tasks like summarization and translation.
    
    \item \textbf{ViT5} \cite{phan2022vit5}: A text-to-text transformer based on T5, trained on a large Vietnamese corpus for both generative and classification tasks.

    \item \textbf{ViSoBERT} \cite{nguyen_visobert_2023}: Designed for noisy and informal social media text, including teencode, emojis, and abbreviations.

    \item \textbf{CafeBERT} \cite{do-etal-2024-vlue}: Optimized for conversational and user-generated Vietnamese content, capturing informal and contextual expressions.
    
\end{itemize}

\section{Empirical Results}
\label{sec:results}


\subsection{Corpus Preparation}

The dataset was divided into training, development, and test sets in an 8:1:1 ratio. Before the corpus was fed into the models, it underwent thorough preprocessing to ensure consistency and quality. Common patterns, like elongated punctuation, were simplified (e.g., ":))))" → ":))"), and duplicated characters and emojis were normalized (e.g., "cườiiiiii" → "cười"; \includesvg[scale=0.015]{emojis/grinning-face-with-smiling-eyes}\includesvg[scale=0.015]{emojis/grinning-face-with-smiling-eyes} → \includesvg[scale=0.015]{emojis/grinning-face-with-smiling-eyes}). In addition, emoji preprocessing, informal expressions, and teencode were optionally normalized depending on the experimental scenario.

\subsection{Experimental Settings}
The experiments were designed to evaluate the impact of different preprocessing strategies and dataset configurations on model performance. Two emoji preprocessing scenarios and one textual normalization strategy were applied:
\begin{itemize}
\item \textbf{Scenario 1} retains emojis in their original form and applies rule-based normalization for teencode and informal words using a manually curated dictionary.
\item \textbf{Scenario 2} converts emojis to Vietnamese textual description (e.g., \includesvg[scale=0.015]{emojis/crying-face} → "khóc"), while also applying the same rule-based normalization for teencode and misspellings.
\item \textbf{Scenario 3} replaces the manual normalization step with ViSoLex \cite{nguyen_visolex_2025}, a model-based lexical normalization system. ViSoLex is instantiated using ViSoBERT and trained to predict standard forms of non-standard words (NSWs) in Vietnamese social media text (e.g, "đung v" → "Đúng vậy."). ViSoLex can be instantiated with either BARTpho or ViSoBERT as the backbone model; in this study, we adopt the ViSoBERT-based variant using a publicly available pretrained checkpoint. This scenario does not alter emojis but uses a learning-based method to normalize textual noise, including teencode and misspellings.

\end{itemize}
Eight selected models were trained under all three settings, and results were compared to determine the optimal preprocessing and normalization strategy for the dataset. To enhance model robustness and mitigate overfitting, a Dropout layer with a dropout rate of $p=0.2$ was appended to randomly deactivate 20\% of the nodes during training. Additionally, a fully connected (FC) layer was added, with the number of output nodes corresponding to the 28 target labels.

The models were trained for $12$ epochs using the AdamW optimizer with an initial learning rate of $5\times 10^{-5}$. A linear learning rate scheduler was employed, in which the learning rate gradually increased during the warmup phase (the 1st epoch) and then linearly decreased until the end of training, approaching zero. For the loss function, Binary Cross-Entropy with Logits was adopted to handle the multilabel classification setting, where each label is predicted independently. To mitigate class imbalance, a positive class weight $\texttt{pos\_weight}$ was computed for each label as 
$\texttt{pos\_weight}_i = \frac{N - n_i}{n_i}$,
where $N$ is the total number of training samples and $n_i$ is the number of positive samples for label $i$. This weighting scheme ensures that minority classes receive greater emphasis during training, thereby improving performance on underrepresented emotion labels.

Finally, the evaluation metrics used in this study are the Macro F1-score and the Weighted F1-score, chosen to provide a comprehensive analysis of model performance across both large and small sentiment classes.

\subsection{Experimental Results}

In this study, we compare the performance of eight models — mBERT (cased), XLM-R\textsubscript{Base}, PhoBERT\textsubscript{Base}v2, ViBERT, BARTpho-syllable\textsubscript{Base}, ViT5\textsubscript{Base}, ViSoBERT, and CafeBERT — across three scenarios:
(1) retaining emojis in their original form,
(2) converting emojis to Vietnamese textual descriptions, and 
(3) applying lexical normalization using ViSoLex.
To compare three processing strategies, Macro-F1 (MF1) scores were evaluated on the development set. As shown in Table \ref{tab:dev_results}, Scenario 1 (preserving emojis in original form) consistently outperformed the other two across most models, the highest MF1 was achieved by ViSoBERT under Scenario 1 (62.33\%), followed closely by CafeBERT (61.45\%).
The results suggest that preserving raw emojis, alongside basic text normalization, helps models better capture affective cues in informal social media text. While Scenario 2 yielded modest gains for some models (e.g., PhoBERT and ViT5), it generally underperformed compared to Scenario 1. Scenario 3, which replaced manual lexical corrections with automatic normalization, showed comparable performance but did not surpass targeted manual corrections. One potential reason is that the ViSoLex model used in this study relies on a publicly available checkpoint pretrained on general Vietnamese social media data. Since it was not fine-tuned on the target corpus, its normalization capacity may not fully align with domain-specific noise or stylistic nuances.

\begin{table}[ht!]
\centering
\resizebox{.4\textwidth}{!}{
\begin{tabular}{llr}
\toprule
\textbf{Scenario} & \textbf{Model} & \textbf{MF1 (\%)} \\
\midrule
\multirow{8}{*}{1} 
  & mBERT (cased)             & 50.36 \\
  & XLM-R\textsubscript{Base}              & 56.71 \\
  & PhoBERT\textsubscript{Base}v2          & 57.03 \\
  & viBERT                   & 49.91 \\
  & BARTpho-syllable\textsubscript{Base}    & 49.96 \\
  & ViT5\textsubscript{Base}                & 55.96 \\
  & ViSoBERT                 & \textbf{62.33} \\
  & CafeBERT                 & 61.45 \\
  
\midrule
\multirow{8}{*}{2} 
  & mBERT (cased)             & 53.08 \\
  & XLM-R\textsubscript{Base}              & 56.22 \\
  & PhoBERT\textsubscript{Base}v2          & 60.08 \\
  & viBERT                   & 53.34 \\
  & BARTpho-syllable\textsubscript{Base}    & 53.22 \\
  & ViT5\textsubscript{Base}                & 58.57 \\
  & ViSoBERT                 & 62.01 \\
  & CafeBERT                 & 60.73 \\
  
\midrule
\multirow{8}{*}{3} 
  & mBERT (cased)             & 49.98 \\
  & XLM-R\textsubscript{Base}              & 56.01 \\
  & PhoBERT\textsubscript{Base}v2          & 56.00 \\
  & viBERT                   & 49.75 \\
  & BARTpho-syllable\textsubscript{Base}    & 49.66 \\
  & ViT5\textsubscript{Base}                & 55.17 \\
  & ViSoBERT                 & 61.18 \\
  & CafeBERT                 & 59.89 \\
  
\bottomrule
\end{tabular}
}
\caption{Macro F1 Scores (\%) on Development Set.}
\label{tab:dev_results}
\end{table}

Based on development results, Scenario 1 was selected for final evaluation on the test set. As shown in Table \ref{tab:test_results}, ViSoBERT again achieved the highest performance with an MF1 of  61.50\% and a WF1 (Weighted F1) of 63.26\%, confirming its effectiveness in handling informal, emoji-rich Vietnamese social media text.

\begin{table}[h!]
    \centering
    \resizebox{.4\textwidth}{!}{
    \begin{tabular}{lrr}
    \toprule
    \textbf{Model} & \textbf{MF1 (\%)} & \textbf{WF1 (\%)} \\
    \midrule
    mBERT (cased)                 & 51.54 & 53.25 \\
    XLM-R\textsubscript{Base}     & 56.21 & 58.15 \\
    PhoBERT\textsubscript{Base}v2              & 56.33 & 57.94 \\
    viBERT                       & 51.41 & 52.76 \\
    BARTpho-syllable\textsubscript{Base} & 51.49 & 53.38 \\
    ViT5\textsubscript{Base}     & 55.92 & 57.54 \\
    CafeBERT                     & 61.17 & 62.74 \\
    ViSoBERT                     & \textbf{61.50} & \textbf{63.26} \\
    \bottomrule
    \end{tabular}
    }
    \caption{Performance on Test Set under Scenario 1.}
    \label{tab:test_results}
\end{table}

To further understand model behavior, Table \ref{viso_metrics} presents a detailed breakdown of ViSoBERT's precision, recall, and F1-score for each of the 27 emotion categories + Neutral. This analysis provides insights into the model's strengths and weaknesses across both frequent and less common expressions.

\begin{table}[h!]
\centering
\small 
\begin{tabular}{lccc}
\toprule
\textbf{Emotion} & \textbf{Precision} & \textbf{Recall} & \textbf{F1-Score} \\
\midrule
    amusement      & 71.03 & 81.28 & 75.81 \\
    excitement     & 56.84 & 55.10 & 55.96 \\
    joy            & 60.25 & 70.59 & 65.01 \\
    love           & 62.57 & 78.32 & 69.57 \\
    desire         & 45.56 & 51.25 & 48.24 \\
    optimism       & 66.88 & 73.94 & 70.23 \\
    caring         & 63.69 & 71.33 & 67.30 \\
    pride          & 73.17 & 69.77 & 71.43 \\
    admiration     & 50.93 & 54.46 & 52.63 \\
    gratitude      & 80.51 & 87.96 & 84.07 \\
    relief         & 49.37 & 65.00 & 56.12 \\
    approval       & 66.10 & 67.83 & 66.95 \\
    realization    & 45.54 & 48.42 & 46.94 \\
    surprise       & 57.95 & 60.00 & 58.96 \\
    curiosity      & 64.42 & 67.00 & 65.69 \\
    confusion      & 47.78 & 51.19 & 49.43 \\
    fear           & 70.30 & 72.45 & 71.36 \\
    nervousness    & 50.53 & 50.00 & 50.26 \\
    remorse        & 73.49 & 70.11 & 71.76 \\
    embarrassment  & 77.31 & 85.98 & 81.42 \\
    disappointment & 47.27 & 52.79 & 49.88 \\
    sadness        & 65.82 & 82.61 & 73.26 \\
    grief          & 58.04 & 75.58 & 65.66 \\
    disgust        & 48.73 & 54.61 & 51.51 \\
    anger          & 60.20 & 62.43 & 61.30 \\
    annoyance      & 58.18 & 65.31 & 61.54 \\
    disapproval    & 36.79 & 34.82 & 35.78 \\
    neutral        & 45.05 & 43.10 & 44.05 \\
    \midrule
    macro avg      & 59.08 & 64.40 & 61.50 \\
    weighted avg   & 60.23 & 66.84 & 63.26 \\
\bottomrule
\end{tabular}
\caption{Precision, Recall, and F1-Score per Emotion Category.}
\label{viso_metrics}
\end{table}

\subsection{Result Analysis}
Experimental results indicate that emoji preprocessing plays a critical role in sentiment classification tasks involving Vietnamese social media content. Across all models, retaining emojis in their original form (Scenario 1) consistently yielded better performance across models. This indicates that raw emojis, when combined with basic manual normalization, help models better capture affective and contextual cues commonly found in informal text. The effect of text pre-processing on three scenarios by ViSoBERT is illustrated through several examples in Table \ref{tab:scenarios_predictions} from the Appendix \ref{appendix_effect}. 

Among the evaluated models, ViSoBERT demonstrated the strongest performance, achieving the highest scores on both the development and test sets. Its pre-training on informal and noisy Vietnamese text, including emoji-rich content from social media platforms, gives it a notable advantage in this domain. Particularly, CafeBERT in Scenario 1 also performed competitively, showing its effectiveness in processing user-generated and conversational text.

\begin{table*}
    \centering
    \renewcommand{\arraystretch}{1.2}
    \resizebox{.75\textwidth}{!}{
    \begin{tabular}{|c|ccc|ccccc|c|}
    \hline
    \multirow{2}{*}{\textbf{GT count}} 
      & \multicolumn{3}{c|}{\textbf{Predicted vs GT (\%)}} 
      & \multicolumn{5}{c|}{\textbf{Correct labels (\%)}} 
      & \multirow{2}{*}{\shortstack{\textbf{F1-macro} \\ (\%)}} \\
    \cline{2-9}
      & Exact & Fewer & More & 1 & 2 & 3 & 4 & 5 &  \\
    \hline
    1 & 50.37 & 2.36  & 47.28 & 74.23 &   -    &    -   &     -  &   -   & 52.56 \\
    2 & 48.62 & 21.29 & 30.08 & 43.01 & 45.66 &    -   &     -  &   -   & 61.75 \\
    3 & 37.37 & 46.13 & 16.49 & 21.39 & 46.39 & 23.71 &     -  &    -  & 61.18 \\
    4 & 27.08 & 64.58 & 8.33  & 10.42 & 31.25 & 41.67 & 10.42 &  -    & 59.09 \\
    5 & 14.29 & 85.71 & 0.00  & 14.29 & 14.29 & 28.57 & 42.86 & 0.00 & 39.35 \\
    \hline
    \end{tabular}
    }
    \caption{Performance breakdown of ViSoBERT on the development set (Scenario 1).}
    \label{tab:correct-failure-statistics}
\end{table*}

Analysis of the per-category performance in Table \ref{viso_metrics} reveals that ViSoBERT performs particularly well in high-frequency and expressive emotions such as gratitude, embarrassment, amusement, and sadness. However, lower F1-scores were observed in more ambiguous or less frequent categories like disapproval, neutral, and realization, suggesting that these emotions are more challenging to detect and may benefit from additional data or targeted augmentation strategies.

To further understand the linguistic characteristics of different sentiment regions, we analyze frequent trigram patterns of word distribution based on four groups: positive sentiment, ambiguous sentiment, negative sentiment, and mixed sentiment. We found that positive sentiment regions often contain action or praise-oriented emotion. In contrast, negative sentiment features strong negation or factual statements. Ambiguous regions include questioning or hedging expressions, while mixed sentiment shows emotions span two or more sentiment groups. The varied patterns in this chart reflect the coexistence of conflicting emotional cues, underscoring the challenges of modeling multi-label sentiment classification. To better understand the sentiment distribution, the Sunburst charts illustrating the most frequent trigram patterns are shown in Figure \ref{fig:sunburst_sentiment} from the Appendix \ref{appendix_ling}. In addition, we note that some common informal Vietnamese words appear across multiple emotion categories. A more detailed discussion together with the representative frequent words for each of the 27 categories is provided in Table \ref{tab:top5-words} from the Appendix \ref{appendix_topword}.


\subsection{Error Analysis}

To provide an in-depth analysis of ViSoBERT's performance across samples with varying ground-truth (GT) label counts, Table \ref{tab:correct-failure-statistics} analyzes ViSoBERT's performance across samples with different ground-truth (GT) label counts. For instance, when GT = 2, 48.62\% of samples had exactly two predicted labels, 21.29\% fewer, and 30.08\% more. In the same row, the model predicted both labels correctly in 45.66\% of cases and only one in 43.01\%. The corresponding F1 score was 61.75\%. Overall, the model performs best on samples with GT = 2, while showing reasonably good performance on low to moderately complex multi-label cases (GT = 1-4). However, it suffers from both label quantity mismatch and semantic under-coverage as the number of ground-truth labels increases. More explanation about Table \ref{tab:correct-failure-statistics} can be found in the Appendix \ref{appendix_scenario1}.

In our error analysis, we focus primarily on emotion categories with low F1-scores, which indicate difficulties in both precision and recall. To gain a clearer understanding of the model's performance across individual emotions, Table \ref{tab:misclassified-labels} in the Appendix \ref{appendix_missclassified} lists pairs of emotion labels that are frequently confused with one another.

Besides, the quantities of FP and FN for each of the 28 category labels achieved by ViSoBERT in the development set are shown in Figure \ref{fig:confusion_fp_fn} in the Appendix \ref{appendix_missclassified}.
More details about misclassification word by emotion category can be found in Table \ref{tab:misclass-words} from the Appendix \ref{appendix_ambigous}. 
\section{Conclusion}
\label{sec:conclusion}
This paper presents ViGoEmotions - a benchmark dataset for fine-grained emotion detection on Vietnamese text, which leverages LLMs for annotation assistance under human supervision. The dataset contains 20,664 comments with about 27 emotion labels, capturing the linguistic diversity and informal nature of user-generated content. Besides, we evaluate the performance of various robust BERT classification models for fine-grained emotion detection based on the ViGoEmotions. We also investigate three preprocessing scenarios for the task: (1) retaining emojis in their original form, (2) converting emojis into Vietnamese textual descriptions, and (3) applying text normalization using ViSoLex. The ViSoBERT running with the retaining emojis scenario achieved the highest performance in both Macro F1 and Weighted F1 scores, which are 61.50\% and 63.26\%, respectively. These findings highlight the importance of integrating informal language handling with suitable emoji presentation in emotion classification and demonstrate the value of adapting pre-trained models to the characteristics of social media text. On the other hand, the semantic overlap and ambiguous contextual cues in linguistics challenge the model in fine-grained emotion detection. Last but not least, the ViGoEmotions provides a foundation for future research in Vietnamese sentiment analysis and emotion detection, particularly in noisy and informal communication contexts. The source codes and ViGoEmotions dataset are publicly available at: \url{https://github.com/ricardo-tran/ViGoEmotions}.

\section*{Limitations}
\label{limitations}
While the proposed dataset and experimental framework offer valuable insights into Vietnamese emotion classification on social media, several limitations remain. First, the dataset, although diverse, is sourced from a limited number of platforms and may not fully capture the breadth of expressions across different social media contexts. Second, the annotation process, despite efforts to ensure consistency, is inherently subjective, particularly for fine-grained emotions with overlapping semantics.

Additionally, the models used in this study are limited by the capacity of current transformer-based architectures and may struggle with subtle contextual cues or sarcasm, which are prevalent in informal online communication. 
Finally, while the exploration of three preprocessing strategies — emoji preservation, emoji-to-text conversion, and lexical normalization via ViSoLex — has shown clear benefits, each remains limited in expressiveness. Retraining or domain-adaptive fine-tuning of the lexical normalizer could potentially enhance performance when adopting ViSoLex. Future work could investigate richer emoji embeddings, context-aware normalization, and multimodal approaches to better capture nuanced emotional expression in informal social media text.
\section*{Acknowledgments}
This research was supported by The VNUHCM-University of Information Technology's Scientific Research Support Fund.

\bibliography{references}
\appendix
\onecolumn
\section{Data Analysis}
\label{sec:appendix}

\subsection{Translation and emotion annotation of Vietnamese sentence from proposed corpus}
\label{appendix_sample}
\begin{table*}[ht!]
    \centering
    \resizebox{.85\textwidth}{!}{
    \begin{tabular}{|c|p{5cm}|p{5cm}|c|}
    
    \hline
    
    \textbf{No.} & \textbf{Vietnamese sentences} & \textbf{English translation} & \textbf{Emotion} \\ \hline
    1 & cần tìm gấp bạn trai đáng yêu như anh này trời đất ơi =)) & I urgently need to find a cute boyfriend like this guy =)) & amusement, desire \\ \hline
    2 & hóng mãi cuối cùng bị cho leo cây \includesvg[scale=0.015]{emojis/unamused-face} & Been waiting forever, only to be stood up in the end \includesvg[scale=0.015]{emojis/unamused-face} & disappointment, annoyance \\ \hline
    3 & khi nào tao thấy thích . mà thích rồi chưa có người thích cùng . & whenever I feel like it. But I like it and no one likes it. & disappointment, sadness \\ \hline
    4 & tội cả con quá \includesvg[scale=0.015]{emojis/loudly-crying-face}\includesvg[scale=0.015]{emojis/loudly-crying-face} một lũ vô nhân tính đã hủy hoại những mầm non tương lai . & Poor child \includesvg[scale=0.015]{emojis/loudly-crying-face}\includesvg[scale=0.015]{emojis/loudly-crying-face} A group of heartless people has destroyed the future seedlings. & sadness, grief, anger \\ \hline
    5 & sao em này lại xinh vậy ? & Why is this one so beautiful? & admiration, surprise \\ \hline
    6 & Không tử tế thì cũng đừng làm khổ người khác chứ \includesvg[scale=0.018]{emojis/weary-face} & If you can't be kind, at least don't make others suffer \includesvg[scale=0.018]{emojis/weary-face} & annoyance, disapproval \\ \hline
    \end{tabular}
    }
    \caption{Translation and emotion annotation of Vietnamese sentences.}
    \label{tab:example_more}
\end{table*}

\subsection{Cohen's Kappa Values}
\label{appendix_kappa}
The Kappa scores presented in Table \ref{tab:kappa_scores} provide an evaluation of the agreement levels for each emotion label. The rows in the Table are sorted by Cohen's Kappa (highest at the top). The scores highlight varying levels of consistency among annotators for different emotions.
All emotion labels achieved Cohen’s Kappa scores above 0.33, indicating at least fair agreement. The highest agreement was observed for fear (0.71), gratitude (0.69), and pride (0.68), suggesting strong annotator consistency for these emotions.

\begin{table}[ht!]
    \centering
    \resizebox{.3\textwidth}{!}{
    \begin{tabular}{l r}
    \toprule
    \textbf{Label} & \textbf{Kappa Score} \\
    \midrule

    fear            & \textbf{0.7148} \\
    gratitude       & \textbf{0.6911} \\
    pride           & \textbf{0.6849} \\
    embarrassment   & 0.6402 \\
    amusement       & 0.6105 \\
    optimism        & 0.6045 \\
    desire          & 0.6028 \\
    curiosity       & 0.6005 \\
    love            & 0.5979 \\
    anger           & 0.5959 \\
    sadness         & 0.5786 \\
    surprise        & 0.5720 \\
    realization     & 0.5672 \\
    confusion       & 0.5591 \\
    approval        & 0.5454 \\
    joy             & 0.5444 \\
    relief          & 0.5441 \\
    caring          & 0.5421 \\
    admiration      & 0.5399 \\
    grief           & 0.5379 \\
    remorse         & 0.5375 \\
    excitement      & 0.5320 \\
    annoyance       & 0.5139 \\
    disgust         & 0.4908 \\
    nervousness     & 0.4920 \\
    neutral         & 0.4583 \\
    disappointment  & 0.4250 \\
    disapproval     & \textit{0.3400} \\
    \bottomrule
    \end{tabular}
    }
\caption{Kappa Scores by Emotion Label.}
\label{tab:kappa_scores}
\end{table}

\subsection{Correlation between emotions}
\label{appendix_correation}
\begin{figure}[ht!]
    \centering
    \includegraphics[width=\linewidth]{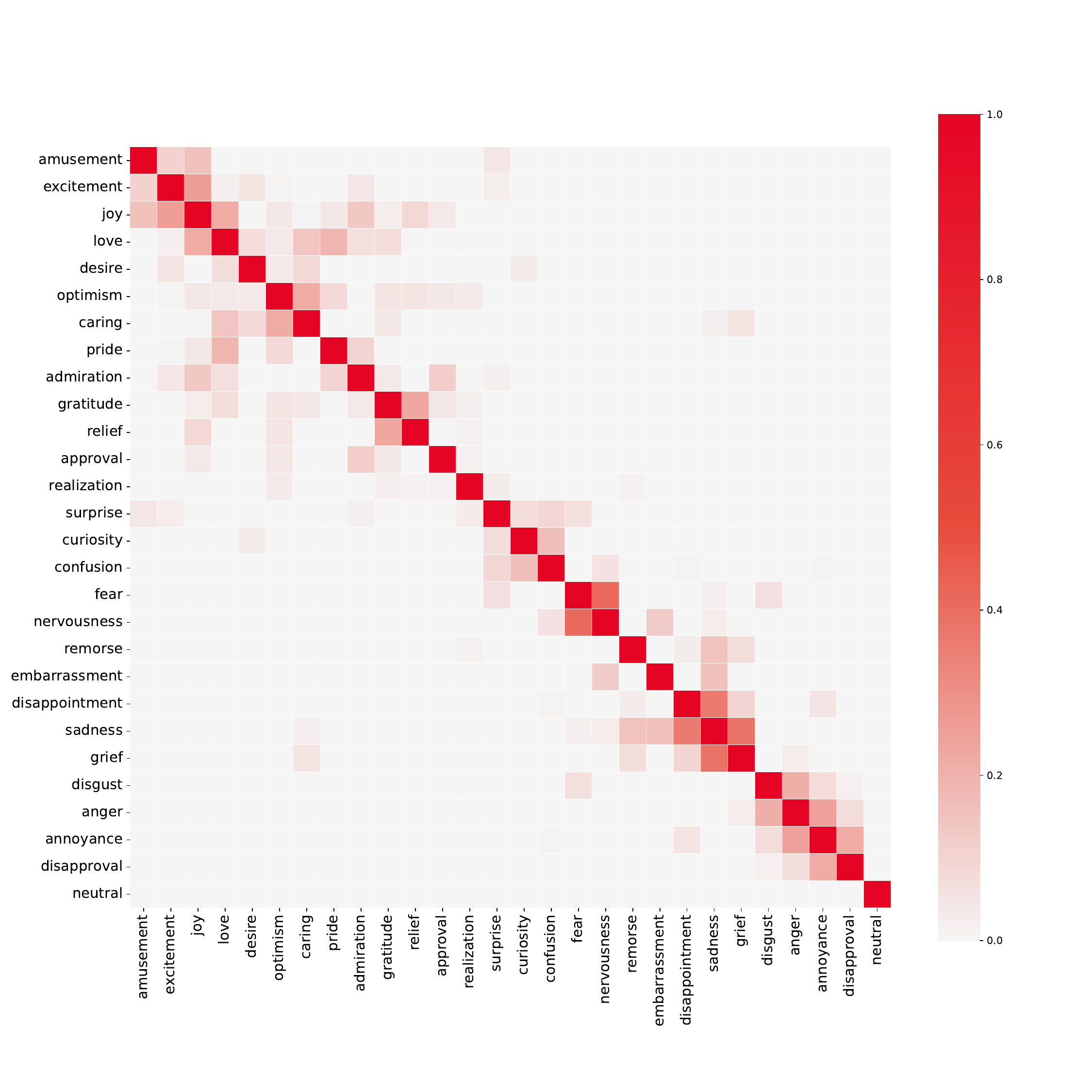}
    \caption{Pearson Correlation Heatmap of Emotion Co-occurrences.}
    \label{fig:pearson_correlation}
\end{figure}

Figure \ref{fig:pearson_correlation} presents the complete correlation heatmap, where warmer colors indicate stronger positive correlations. Additionally, Table \ref{tab:emotion_correlation} shows the top-5 most correlated for each emotion. 

\begin{table}[ht!]
    \centering
    \begin{tabular}{>{\raggedright\arraybackslash}l >{\raggedright\arraybackslash}l}
    \toprule
    \textbf{Emotion} & \textbf{Top-5 Correlated Emotions} \\
    \midrule
    amusement & joy, excitement, surprise, relief, admiration \\
    excitement & joy, amusement, desire, admiration, surprise \\
    joy & excitement, love, amusement, admiration, relief \\
    love & joy, pride, caring, gratitude, desire \\
    desire & caring, love, excitement, optimism, curiosity \\
    optimism & caring, pride, relief, gratitude, joy \\
    caring & optimism, love, desire, grief, gratitude \\
    pride & love, admiration, optimism, joy, excitement \\
    admiration & joy, approval, pride, love, excitement \\
    gratitude & relief, love, optimism, caring, approval \\
    relief & gratitude, joy, optimism, realization, caring \\
    approval & admiration, gratitude, optimism, joy, realization \\
    realization & optimism, surprise, gratitude, approval, relief \\
    surprise & confusion, curiosity, fear, amusement, realization \\
    curiosity & confusion, surprise, desire, caring, nervousness \\
    confusion & curiosity, surprise, nervousness, disappointment, annoyance \\
    fear & nervousness, surprise, disgust, sadness, embarrassment \\
    nervousness & fear, embarrassment, confusion, sadness, disappointment \\
    remorse & sadness, grief, disappointment, realization, desire \\
    embarrassment & sadness, nervousness, fear, disappointment, remorse \\
    disappointment & sadness, grief, annoyance, remorse, confusion \\
    sadness & grief, disappointment, embarrassment, remorse \\
    grief & sadness, disappointment, remorse, caring, anger \\
    disgust & anger, annoyance, fear, disapproval, grief \\
    anger & annoyance, disgust, disapproval, grief, fear \\
    annoyance & anger, disapproval, disgust, disappointment, confusion \\
    disapproval & annoyance, anger, disgust, confusion, fear \\
    neutral & realization, relief, confusion, disapproval, curiosity \\
    \bottomrule
    \end{tabular}%
\caption{Top-5 Correlated Emotions (from Pearson Correlation Matrix)}
\label{tab:emotion_correlation}
\end{table}

\subsection{Linguistic patterns}
\label{appendix_ling}
\begin{figure}[ht!]
    \centering
    \begin{subfigure}{0.48\textwidth}
        \includegraphics[width=\textwidth]{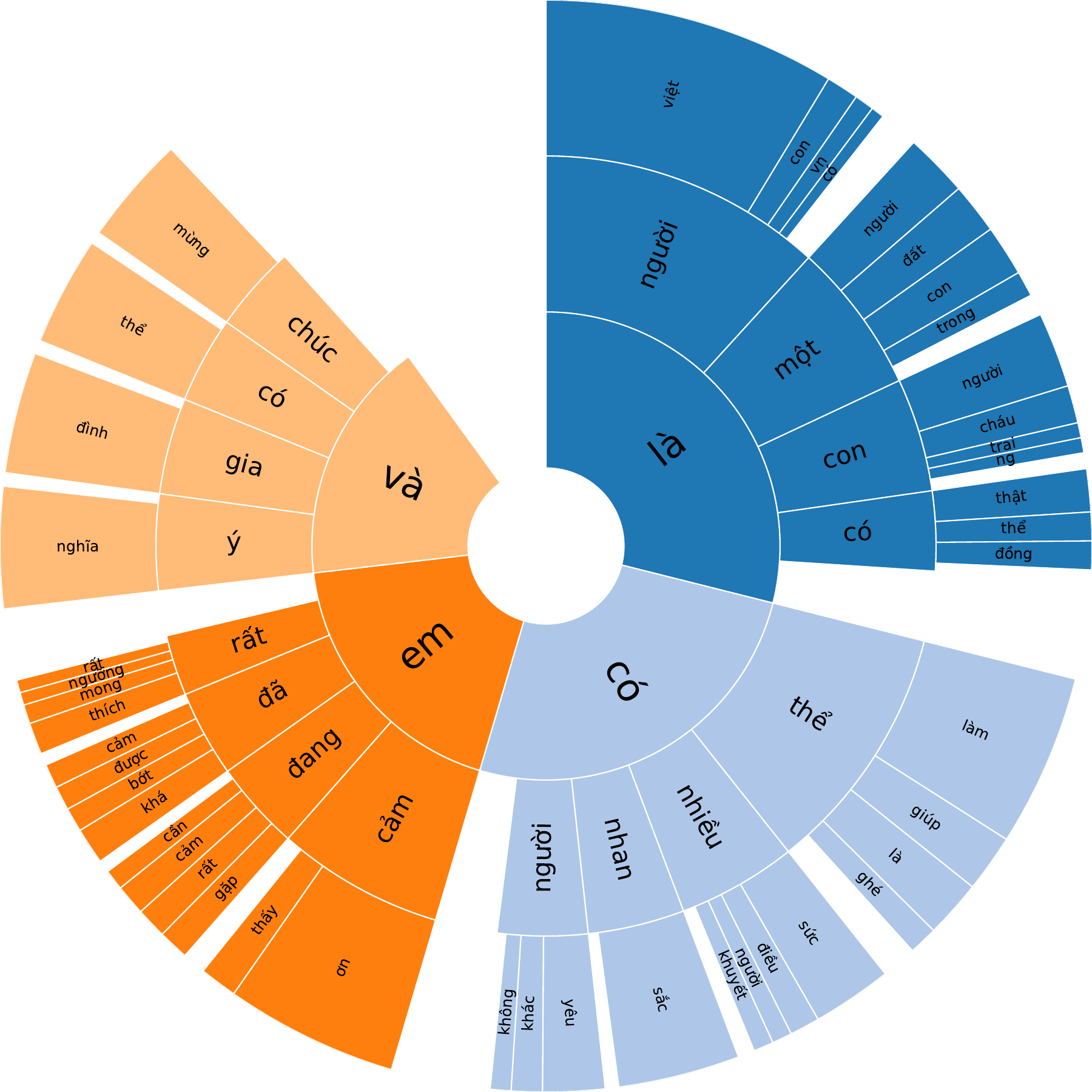}
        \caption{Positive sentiment}
        \label{fig:sunburst_positive}
    \end{subfigure}
    \hfill
    \begin{subfigure}{0.48\textwidth}
        \includegraphics[width=\textwidth]{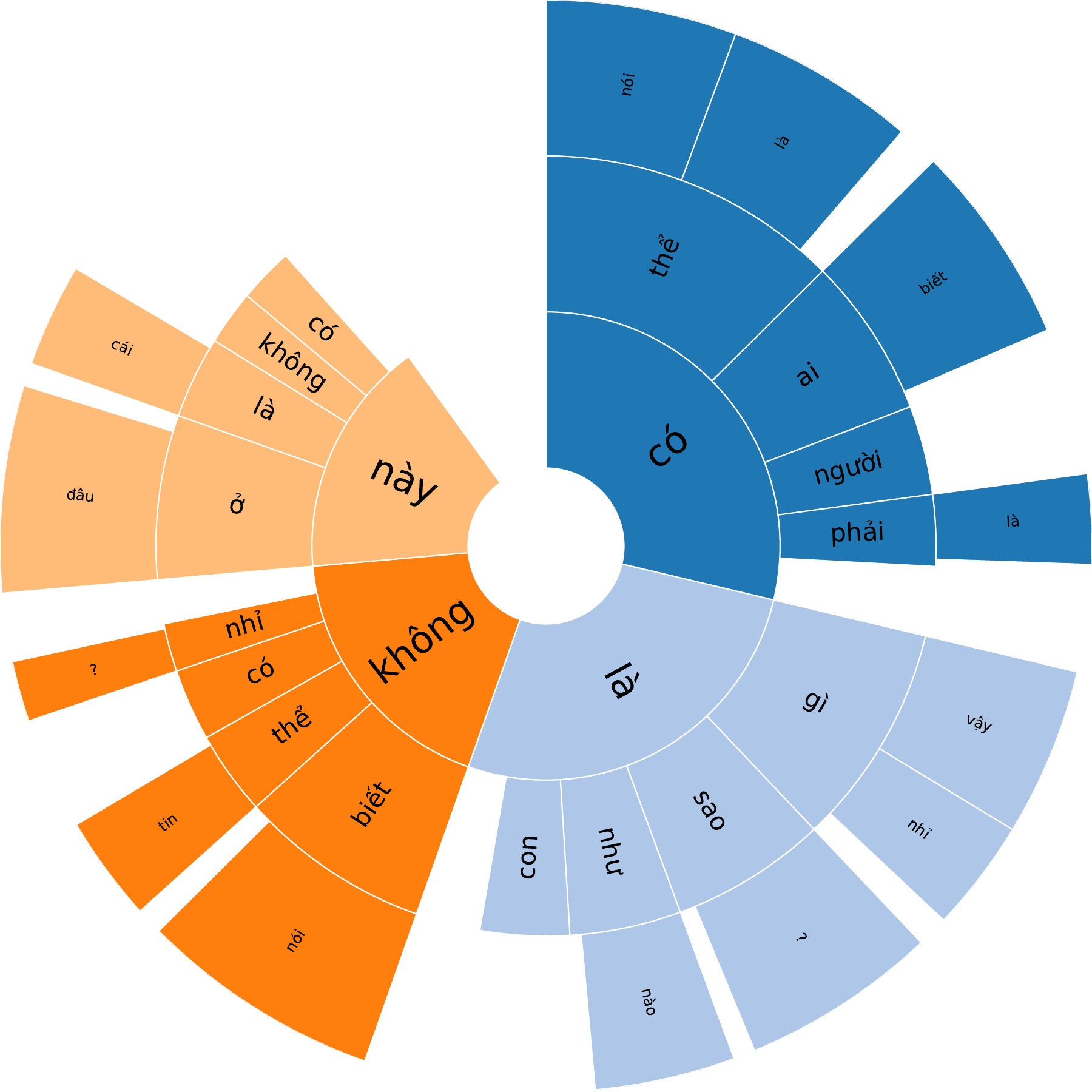}
        \caption{Ambiguous sentiment}
        \label{fig:sunburst_ambiguous}
    \end{subfigure}
    \hfill

    \begin{subfigure}{0.48\textwidth}
        \includegraphics[width=\textwidth]{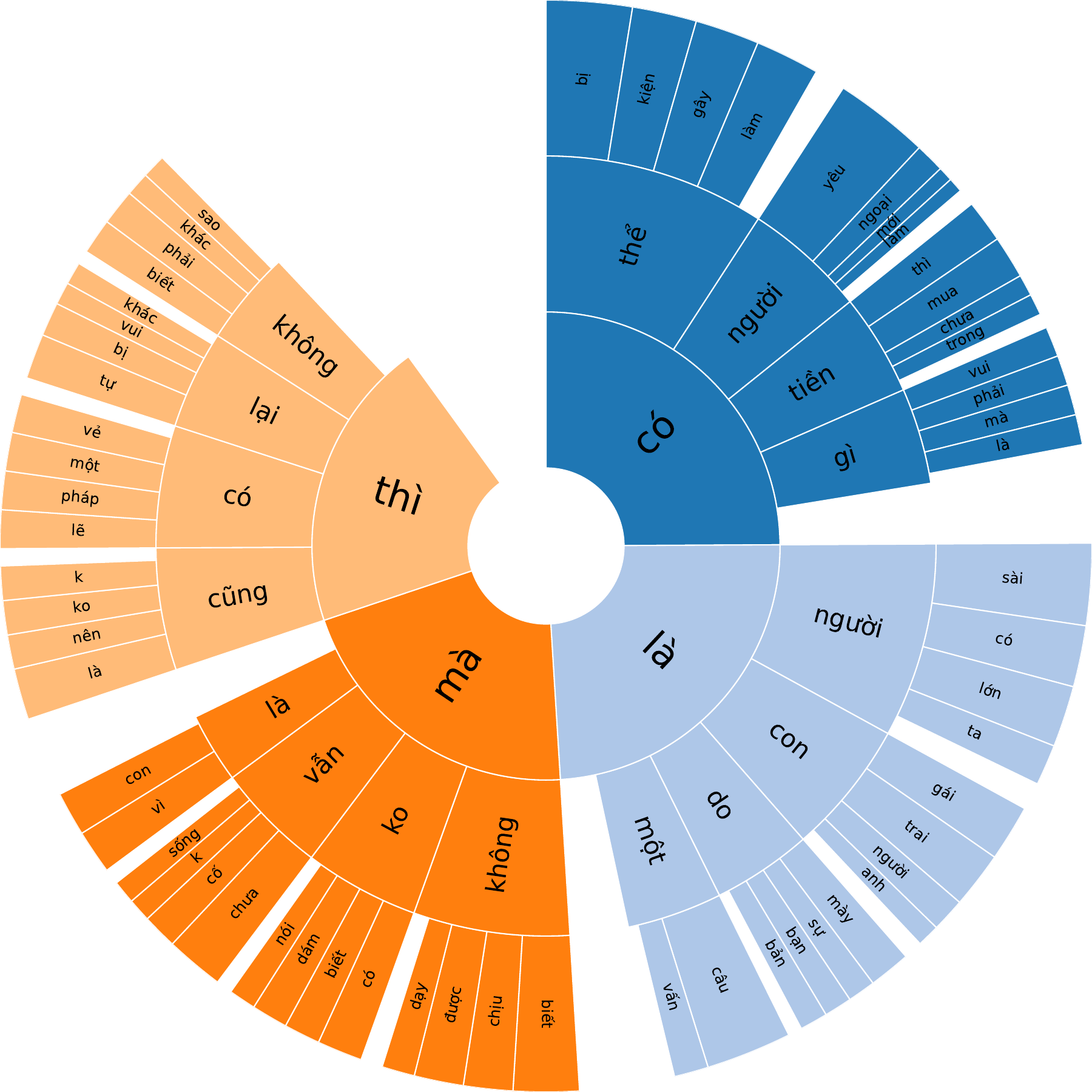}
        \caption{Negative sentiment}
        \label{fig:sunburst_negative}
    \end{subfigure}
    \hfill
    \begin{subfigure}{0.48\textwidth}
        \includegraphics[width=\textwidth]{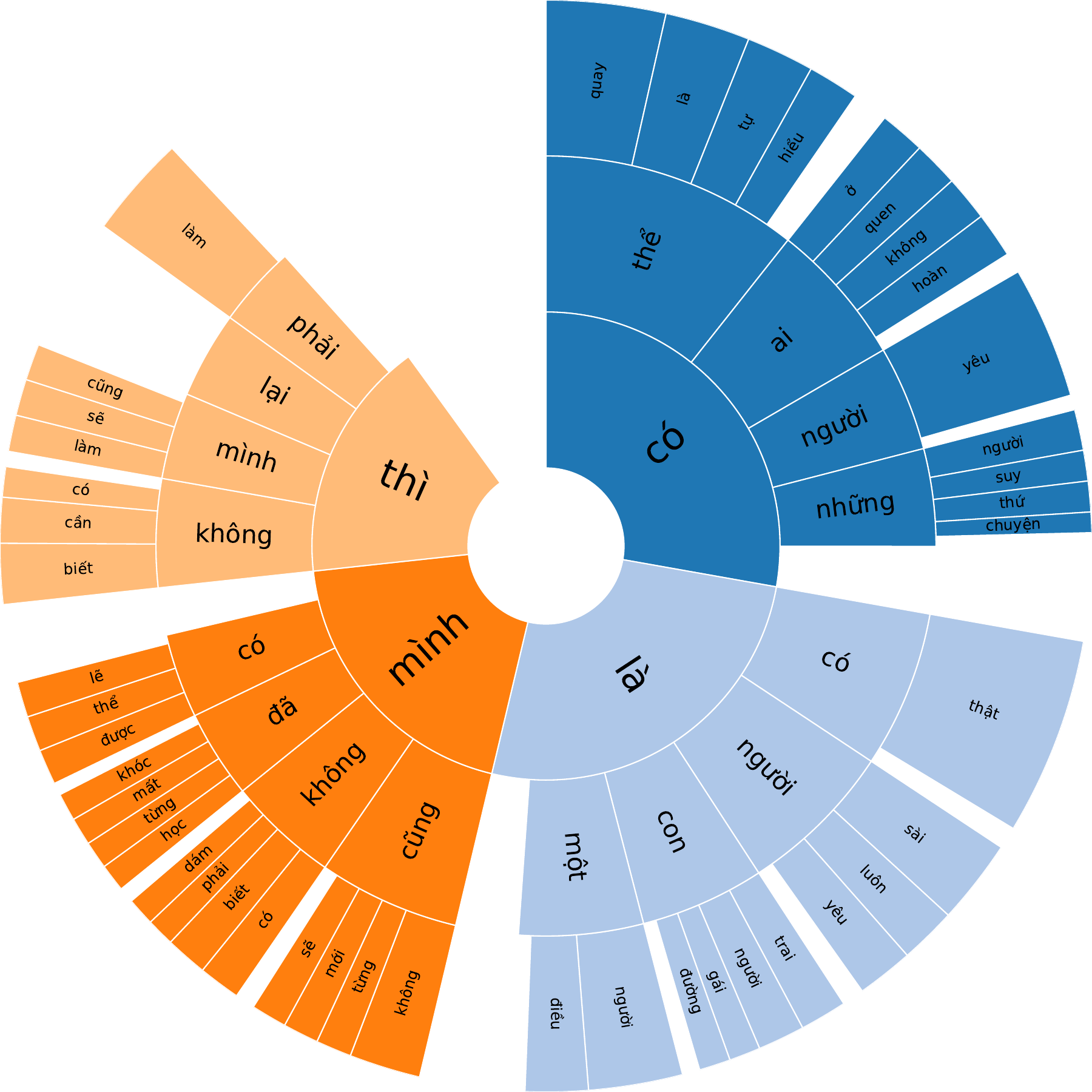}
        \caption{Mixed sentiment}
        \label{fig:sunburst_mixed}
    \end{subfigure}
    \hfill

    \caption{Sunburst chart of 3-gram (trigram).}
    \label{fig:sunburst_sentiment}
\end{figure}

The linguistic patterns in the corpus are presented in trigrams. A sunburst chart (Figure \ref{fig:sunburst_sentiment}) provides a hierarchical view of the most frequent trigrams in the training set.


\subsection{Top words for each emotion}
\label{appendix_topword}
The wordcloud visualization displays the most prominent words associated with each emotional category in the dataset. From Table \ref{tab:top5-words}, we identified the top five most representative words for each emotion category, including \colorbox{greenhead}{positive}, \colorbox{yellowhead}{ambiguous}, \colorbox{redhead}{negative} in the training set.
The table presents the five most salient words for each emotion, ranked by their normalized scores (in parentheses). These scores reflect the relative importance of each word within its respective emotion category. 
We observe that some words, such as \textit{đi}, \textit{ko}, \textit{t}, \textit{tao}, appear across multiple emotion categories. This reflects characteristics of informal Vietnamese on social media, where frequent use of personal pronouns and common verbs is pervasive regardless of sentiment. These words often act as linguistic carriers or discourse markers rather than explicit emotional indicators.

{\scriptsize 
\begin{longtable}{|>{\centering\arraybackslash}p{1.35cm}
                  |>{\centering\arraybackslash}p{1.35cm}
                  |>{\centering\arraybackslash}p{1.35cm}
                  |>{\centering\arraybackslash}p{1.35cm}
                  |>{\centering\arraybackslash}p{1.35cm}
                  |>{\centering\arraybackslash}p{1.35cm}
                  |>{\centering\arraybackslash}p{1.35cm}
                  |>{\centering\arraybackslash}p{1.35cm}
                  |>{\centering\arraybackslash}p{1.35cm}|}

\hline
\cellcolor{greenhead}\textbf{amusement} & 
\cellcolor{greenhead}\textbf{love} & 
\cellcolor{greenhead}\textbf{caring} & 
\cellcolor{greenhead}\textbf{gratitude} & 
\cellcolor{yellowhead}\textbf{realization} & 
\cellcolor{yellowhead}\textbf{confusion} & 
\cellcolor{redhead}\textbf{remorse} & 
\cellcolor{redhead}\textbf{sadness} & 
\cellcolor{redhead}\textbf{anger} \\
\hline
tao (1.0) & yêu (1.0) & thương (1.0) & cảm ơn (1.0) & ta (1.0) & ko (1.0) & hối hận (1.0) & đi (1.0) & đi (1.0) \\
đi (.92) & thương (.92) & đi (.74) & video (.37) & ko (.92) & t (.5) & ko (.44) & ko (.86) & tao (.8) \\
t (.49) & Việt Nam (.7) & chúc (.7) & e (.3) & học (.87) & đi (.47) & đi (.38) & t (.72) & thằng (.52) \\
vãi (.37) & đi (.29) & ko (.62) & ạ (.19) & đi (.82) & tao (.43) & thân (.3) & tao (.68) & mày (.49) \\
cười (.32) & tao (.29) & sống (.54) & chia sẻ (.18) & thân (.53) & học (.31) & thể (.25) & thương (.51) & đi (.45) \\
\hline
\cellcolor{greenhead}\textbf{excitement} & 
\cellcolor{greenhead}\textbf{desire} & 
\cellcolor{greenhead}\textbf{pride} & 
\cellcolor{greenhead}\textbf{relief} & 
\cellcolor{yellowhead}\textbf{surprise} & 
\cellcolor{redhead}\textbf{fear} & 
\cellcolor{redhead}\textbf{embarrassment} & 
\cellcolor{redhead}\textbf{grief} & 
\cellcolor{redhead}\textbf{annoyance} \\
\hline
đi (1.0) & mong (1.0) & Việt Nam (1.0) & cảm ơn (1.0) & tao (1.0) & sợ (1.0) & ngại (1.0) & đi (1.0) & đi (1.0) \\
tao (.75) & đi (.89) & hào (.55) & video (.88) & trời (.88) & đi (.41) & ti (.88) & nội (.91) & tao (.82) \\
phim (.58) & tao (.51) & hào Việt (.29) & đi (.76) & tưởng (.83) & tao (.33) & t (.87) & thương (.88) & ko (.64) \\
video (.31) & video (.46) & VN (.26) & e (.74) & phim (.81) & t (.25) & ko (.54) & đau (.84) & mày (.44) \\
đỉnh (.3) & e (.45) & ko (.23) & ko (.61) & vãi (.76) & ko (.25) & mặt (.42) & ko (.73) & mấy (.41) \\
\hline
\cellcolor{greenhead}\textbf{joy} & 
\cellcolor{greenhead}\textbf{optimism} & 
\cellcolor{greenhead}\textbf{admiration} & 
\cellcolor{greenhead}\textbf{approval} & 
\cellcolor{yellowhead}\textbf{curiosity} & 
\cellcolor{redhead}\textbf{nervousness} & 
\cellcolor{redhead}\textbf{disappointment} & 
\cellcolor{redhead}\textbf{disgust} & 
\cellcolor{redhead}\textbf{disapproval} \\
\hline
đi (1.0) & đi (1.0) & đẹp (1.0) & chuẩn (1.0) & ko (1.0) & sợ (1.0) & đi (1.0) & đi (1.0) & đi (1.0) \\
tao (.83) & sống (.98) & giỏi (.66) & đi (.67) & e (.73) & đi (.56) & ko (.85) & tao (.67) & ko (.8) \\
phim (.55) & ko (.74) & phim (.6) & lắm (.6) & đi (.71) & ko (.5) & tao (.72) & thằng (.55) & đừng (.47) \\
vui (.48) & ta (.66) & video (.57) & ko (.56) & địa (.49) & t (.39) & t (.53) & mày (.52) & sai (.41) \\
đẹp (.42) & thân (.64) & đi (.57) & cảm ơn (.56) & ạ (.46) & e (.38) & lắm (.42) & ghê (.48) & mấy (.39) \\
\hline
\caption{Top 5 representative words for each emotion category extracted from the wordcloud.}
\label{tab:top5-words} \\
\end{longtable}
}

\subsection{Effect of Preprocessing Scenarios on Predictions}
\label{appendix_effect}
In this part, we examine how different text preprocessing strategies affect emotion classification outcomes. Table \ref{tab:scenarios_predictions} compares ViSoBERT's predicted emotion labels across the three preprocessing scenarios. All predictions are made on the same development set, ensuring the only variable is the preprocessing applied to the input text.
Scenario 1 consistently produces predictions that align most closely with the gold labels. This suggests that minimal preprocessing retains expressive features crucial for ViSoBERT's emotion inference, such as emojis and text normalization...
Scenario 2 introduces semantic drift by replacing informal expressions with formal language (e.g., replacing “cười\_ngại” with generic words), which sometimes alters the emotion being conveyed.
Scenario 3 performs the worst, often stripping away emojis and syntactic quirks that are key emotional indicators. In multiple cases, it leads to empty or incomplete label predictions.

{\footnotesize 
\begin{longtable}{|p{3.5cm}|p{3.5cm}|p{3.5cm}|p{3.5cm}|}
\hline
\textbf{Text, Translation \& Ground Truth} & \textbf{S1's Text \& Prediction} & \textbf{S2's Text \& Prediction} & \textbf{S3's Text \& Prediction} \\
\hline
\endfirsthead

\hline
\textbf{Text, Translation \& Labels} & \textbf{S1's Text \& Predict} & \textbf{S2's Text \& Predict} & \textbf{S3's Text \& Predict} \\
\hline
\endhead

í là xong là phải nhún nhún \textcolor{blue}{zậy} chi \textcolor{blue}{z}\includesvg[scale=0.015]{emojis/grinning-face-with-sweat}\includesvg[scale=0.015]{emojis/grinning-face-with-sweat}\includesvg[scale=0.015]{emojis/grinning-face-with-sweat}\newline  
Once it's done, why do you have to bounce like that?\newline
\textit{[amusement]} &
í là xong là phải nhún nhún \textcolor{red}{vậy} chi \textcolor{red}{vậy}\includesvg[scale=0.015]{emojis/grinning-face-with-sweat} \newline\newline\newline
\textit{[amusement]} &
í là xong là phải nhún nhún \textcolor{red}{vậy} chi \textcolor{red}{vậy cười\_ngại} \newline\newline\newline
\textit{[amusement]} &
í là xong là phải nhún nhún \textcolor{red}{vậy} chi \textcolor{red}{vậy.} \newline\newline\newline
\textit{[confusion, neutral]} \\
\hline

mong bà nội \textcolor{blue}{t} thấy và tự ái\includesvg[scale=0.015]{emojis/relieved} \newline
Hope my grandma sees it and gets offended\newline
\textit{[desire]} &
mong bà nội \textcolor{red}{tao} thấy và tự ái\includesvg[scale=0.015]{emojis/relieved} \newline\newline
\textit{[desire]} &
mong bà nội \textcolor{red}{tao} thấy và tự ái \textcolor{red}{hài\_lòng} \newline\newline
\textit{[love]} &
mong bà nội \textcolor{red}{tôi} thấy và tự ái\includesvg[scale=0.015]{emojis/relieved}\textcolor{red}{.} \newline\newline
\textit{[]} \\
\hline

đúng \textcolor{blue}{v}\includesvg[scale=0.015]{emojis/flushed-face} \newline
That's right\newline
\textit{[approval]} &
đúng \textcolor{red}{vậy}\includesvg[scale=0.015]{emojis/flushed-face} \newline\newline
\textit{[approval]} &
đúng \textcolor{red}{vậy ngượng\_ngùng} \newline\newline
\textit{[approval]} &
đúng \textcolor{red}{vậy\includesvg[scale=0.015]{emojis/flushed-face}.} \newline\newline
\textit{[approval]} \\
\hline

\textcolor{blue}{t} đang thắc mắc là 2 thằng kia đéo hỏi thằng per à \includesvg[scale=0.015]{emojis/thinking-face} \newline
I'm wondering why those two didn't ask @person\newline
\textit{[curiosity, annoyance, confusion]} &
\textcolor{red}{tao} đang thắc mắc là 2 thằng kia đéo hỏi thằng per à \includesvg[scale=0.015]{emojis/thinking-face} \newline\newline\newline
\textit{[curiosity, confusion, annoyance]} &
\textcolor{red}{tao} đang thắc mắc là 2 thằng kia đéo hỏi thằng per à \textcolor{red}{suy\_nghĩ} \newline\newline
\textit{[curiosity, confusion]} &
\textcolor{red}{tôi} đang thắc mắc là 2 thằng kia đéo hỏi thằng per à \includesvg[scale=0.015]{emojis/thinking-face}\textcolor{red}{.} \newline\newline\newline
\textit{[curiosity, confusion]} \\
\hline

định mệnh gớm vãi \includesvg[scale=0.018]{emojis/face-vomiting}\includesvg[scale=0.018]{emojis/face-vomiting}\includesvg[scale=0.018]{emojis/face-vomiting} \newline
So heartbroken\newline
\textit{[disgust, annoyance]} &
định mệnh gớm vãi \includesvg[scale=0.018]{emojis/face-vomiting} \newline\newline
\textit{[disgust, annoyance]} &
định mệnh gớm vãi \textcolor{red}{mắc\_ói} \newline\newline
\textit{[disgust]} &
định mệnh gớm vãi \textcolor{red}{.} \newline\newline
\textit{[disgust]} \\
\hline

lâu rất lâu mới gặp được lần rồi liết mắt nhìn cái \textcolor{blue}{:))))} chưa thấy nó cười \newline
Finally  met after a long time, glanced at each other :)))) haven't seen it smile yet"
\textit{[disappointment, sadness, grief]} &
lâu rất lâu mới gặp được lần rồi liết mắt nhìn cái \textcolor{red}{: ))} chưa thấy nó cười \newline\newline\newline
\textit{[amusement, joy, disappointment, sadness]} &
lâu rất lâu mới gặp được lần rồi liết mắt nhìn cái \textcolor{red}{cười\_lớn} chưa thấy nó cười \newline\newline\newline
\textit{[amusement, joy, love, sadness]} &
lâu rất lâu mới gặp được lần rồi liết mắt nhìn cái \textcolor{red}{: ))} chưa thấy nó cười\textcolor{red}{.} \newline\newline\newline
\textit{[amusement, joy]} \\
\hline

\caption{Comparison of preprocessed input text and predicted emotion labels across three different scenarios. Each scenario represents a variation in text preprocessing strategy applied to the same original utterance.}
\label{tab:scenarios_predictions} \\
\end{longtable}
}

\subsection{Performance breakdown of ViSoBERT on Scenario 1}
\label{appendix_scenario1}
Table \ref{tab:correct-failure-statistics} provides a detailed performance breakdown of the ViSoBERT model under Scenario 1, focusing on how well it handles different levels of multi-label classification difficulty based on the number of ground-truth (GT) labels per input sample. The table analyzes predictions grouped by GT count—the number of true emotion labels associated with each sample in the development set. This count ranges from 1 to 5, reflecting the varying complexity of the data. The meaning of columns in the Table \ref{tab:correct-failure-statistics} is described as belows:

\begin{itemize}
\item The "Predicted vs GT (\%)" column compares the number of predicted labels with the number of gold labels. The "Exact" column captures the proportion of samples where the number of predicted labels matched the GT count, while the "Fewer" and "More" columns correspond to cases where the model predicted fewer or more labels than present in the ground truth, respectively. These three columns sum to 100\% in each row, offering insight into whether the model tends to under- or over-generate labels. For instance, in the row where GT count is 2, 48.62\% of samples received exactly three predicted labels, 21.29\% received fewer, and 30.08\% received more. A strong trend toward under-prediction becomes increasingly evident as the GT count increases, with the model generating fewer labels than needed in the majority of cases for GT counts of 3, 4, and 5.

\item The "Correct labels (\%)" column reflects how many of the GT labels were correctly predicted by the model. Each column in this block corresponds to the percentage of samples for which ViSoBERT correctly predicted exactly 1, 2, 3, 4, or 5 gold labels, with the number of correct predictions constrained by the total number of labels for that sample. For example, in samples where GT count equals 2, ViSoBERT predicted exactly one correct label in 43.01\% of cases, and both labels correctly in 45.66\% of cases. The sum of these percentages is typically less than or equal to 100\%, as some predictions may contain no correct labels at all.

\item The "F1-macro" column reports the macro-averaged F1 score across all emotion labels, computed over all samples with the given GT count. 
\end{itemize}


\subsection{Misclassified among 27 emotion labels}
\label{appendix_missclassified}
Table \ref{tab:misclassified-labels} lists pairs of emotion labels that are frequently confused with one another. These misclassifications often stem from semantic overlap or ambiguous contextual cues. Overall, the word that represents the emotion is mostly ambiguous among different types of emotions that challenge the model in correctly predicting the emotion for a sentence, indicating a linguistic phenomenon in Vietnamese emotion expression language.

\begin{figure}[ht]
    \centering
    \includegraphics[width=\linewidth]{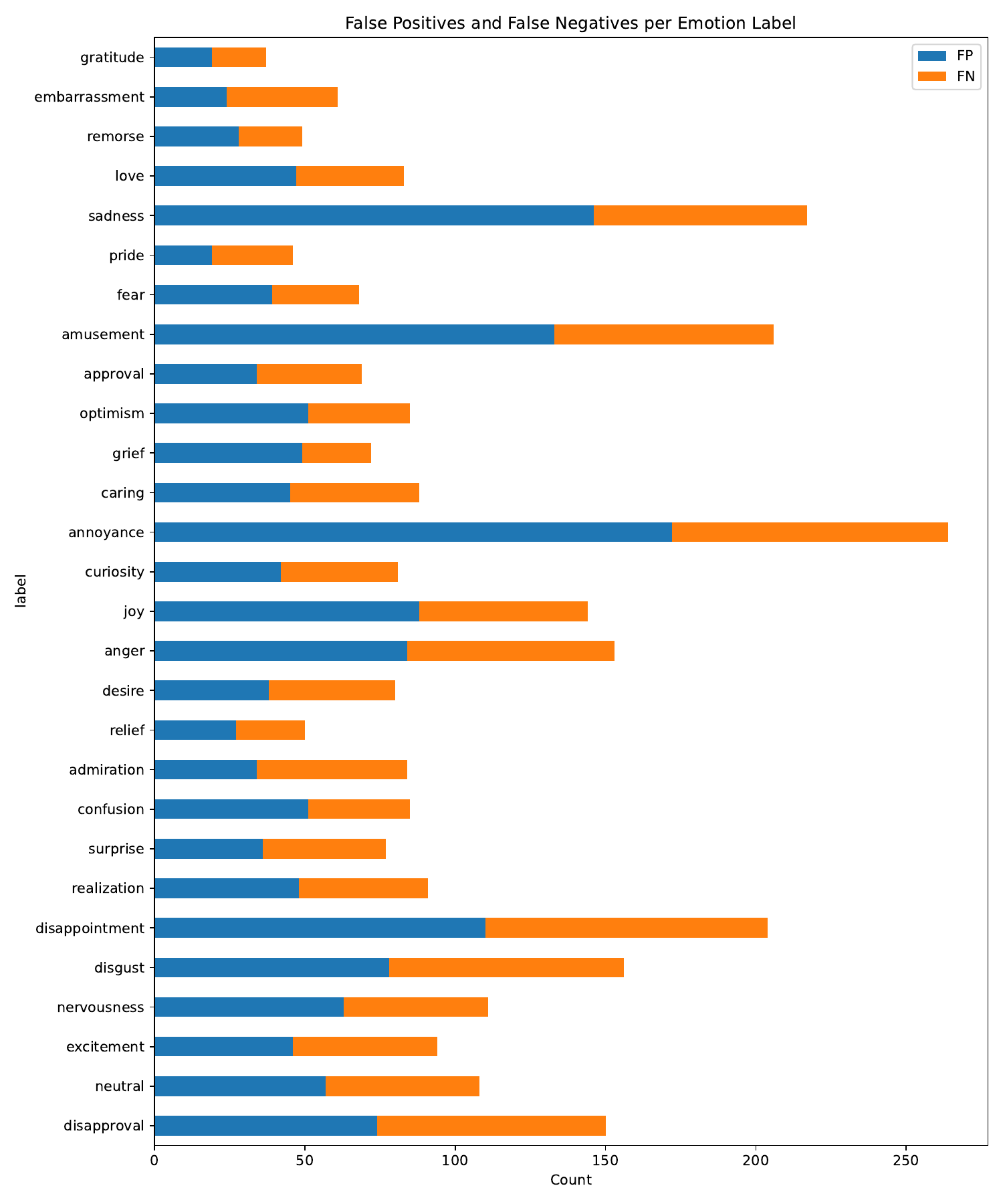}
    \caption{FP and FN for each of the 27 emotions + Neutral.}
    \label{fig:confusion_fp_fn}
\end{figure}

The quantities of FP and FN for each of the 28 category labels achieved by ViSoBERT in the development set are shown in Figure \ref{fig:confusion_fp_fn} from the Appendix.
We found that four emotions, including annoyance, amusement, sadness, and disappointment, are often misclassified. Also, some of the personal pronouns in Vietnamese, like "tao" (I), "mày" (You), and "ta" (We), are likely associated with mispredictions, which could inform future improvements in preprocessing, class weighting, or source-target emotion.

\begin{table}[ht!]
    \footnotesize
    \centering
    \resizebox{.7\textwidth}{!}{
    \begin{tabular}{>{\raggedright\arraybackslash}p{1.9cm} >{\raggedright\arraybackslash}p{5.3cm}}
    \toprule
    \textbf{True Label} & \textbf{Top 3 Misclassified Labels} \\
    \midrule
    amusement & disappointment, annoyance, disgust \\
    excitement & annoyance, sadness, amusement \\
    joy & amusement, sadness, disappointment \\
    love & amusement, optimism, grief \\
    desire & sadness, grief, annoyance \\
    optimism & sadness, approval, admiration \\
    caring & optimism, relief, amusement \\
    pride & annoyance, amusement, disappointment \\
    admiration & amusement, caring, annoyance \\
    gratitude & joy, desire, sadness \\
    relief & sadness, annoyance, realization \\
    approval & realization, optimism, neutral \\
    realization & amusement, disappointment, sadness \\
    surprise & amusement, annoyance, curiosity \\
    curiosity & annoyance, disgust, amusement \\
    confusion & annoyance, sadness, amusement \\
    fear & amusement, annoyance, grief \\
    nervousness & sadness, annoyance, amusement \\
    remorse & annoyance, disappointment, grief \\
    embarrassment & annoyance, sadness, nervousness \\
    disappointment & annoyance, amusement, disapproval \\
    sadness & annoyance, relief, amusement \\
    grief & remorse, fear, disappointment \\
    disgust & annoyance, anger, amusement \\
    anger & annoyance, disapproval, amusement \\
    annoyance & amusement, neutral, sadness \\
    disapproval & anger, annoyance, amusement \\
    neutral & disappointment, annoyance, sadness \\
    \bottomrule
    \end{tabular}
    }
\caption{Top 3 Misclassified Labels per Emotion Classes.}
\label{tab:misclassified-labels}
\end{table}

\subsection{Ambiguous word that causes misclassification for each emotion category}
\label{appendix_ambigous}
Frequent tokens of misclassified samples are shown in Table \ref{tab:misclass-words}.

{\scriptsize
\begin{longtable}{|>{\centering\arraybackslash}p{1.35cm}
                  |>{\centering\arraybackslash}p{1.35cm}
                  |>{\centering\arraybackslash}p{1.35cm}
                  |>{\centering\arraybackslash}p{1.35cm}
                  |>{\centering\arraybackslash}p{1.35cm}
                  |>{\centering\arraybackslash}p{1.35cm}
                  |>{\centering\arraybackslash}p{1.35cm}
                  |>{\centering\arraybackslash}p{1.35cm}
                  |>{\centering\arraybackslash}p{1.35cm}|}

\hline
\cellcolor{greenhead}\textbf{amusement} & 
\cellcolor{greenhead}\textbf{love} & 
\cellcolor{greenhead}\textbf{caring} & 
\cellcolor{greenhead}\textbf{gratitude} & 
\cellcolor{yellowhead}\textbf{realization} & 
\cellcolor{yellowhead}\textbf{confusion} & 
\cellcolor{redhead}\textbf{remorse} & 
\cellcolor{redhead}\textbf{sadness} & 
\cellcolor{redhead}\textbf{anger} \\
\hline
đi (1.0) & chẳng (1.0) & thành (1.0) & lành (1.0) & tui (1.0) & tao (1.0) & chẳng (1.0) & 1 (1.0) & tao (1.0) \\
mày (.57) & tình (.57) & cảm (.75) & tình (1.0) & hôm (1.0) & mày (1.0) & ta (.57) & mày (.75) & đi (.69) \\
tao (.57) & tao (.57) & ơn (.75) & phúc (1.0) & thương (.67) & ghét (.8) & thân (.57) & \includesvg[scale=0.015]{emojis/folded-hands} (.75) & mày (.62) \\
thằng (.57) & thương (.43) & giúp (.75) & video (1.0) & ta (.5) & sinh (.6) & nội (.57) & giải (.75) & mấy (.62) \\
đánh (.57) & 1 (.43) & công (.75) & chủ (1.0) & quen (.5) & bắt (.6) & lỗi (.43) & hàng (.5) & mẹ (.46) \\
\hline
\cellcolor{greenhead}\textbf{excitement} & 
\cellcolor{greenhead}\textbf{desire} & 
\cellcolor{greenhead}\textbf{pride} & 
\cellcolor{greenhead}\textbf{relief} & 
\cellcolor{yellowhead}\textbf{surprise} & 
\cellcolor{redhead}\textbf{fear} & 
\cellcolor{redhead}\textbf{embarassment} & 
\cellcolor{redhead}\textbf{grief} & 
\cellcolor{redhead}\textbf{annoyance} \\
\hline
tao (1.0) & tao (1.0) & haha (1.0) & tui (1.0) & ? (1.0) & tiếng (1.0) & hàng (1.0) & ta (1.0) & mấy (1.0) \\
hai (.75) & mấy (.86) & thong (1.0) & ! (.83) & tao (.75) & tao (1.0) & cảm (1.0) & sợ (.67) & vãi (1.0) \\
thằng (.5) & 1 (.57) & ta (1.0) & buông (.67) & chả (.5) & \includesvg[scale=0.015]{emojis/disappointed-face} (.75) & mấy (1.0) & cố (.67) & học (1.0) \\
vãi (.5) & tui (.43) & khờ (1.0) & cảm (.5) & tưởng (.5) & xong (.75) & trời (1.0) & giới (.33) & bình (1.0) \\
đứa (.5) & bắt (.43) & bos (.5) & động (.5) & 2 (.5) & động (.5) & kéo (1.0) & chiến (.33) & thay (.75) \\
\hline
\cellcolor{greenhead}\textbf{joy} & 
\cellcolor{greenhead}\textbf{optimism} & 
\cellcolor{greenhead}\textbf{admiration} & 
\cellcolor{greenhead}\textbf{approval} & 
\cellcolor{yellowhead}\textbf{curiosity} & 
\cellcolor{redhead}\textbf{nervousness} & 
\cellcolor{redhead}\textbf{disappointment} & 
\cellcolor{redhead}\textbf{disgust} & 
\cellcolor{redhead}\textbf{disapproval} \\
\hline
tao (1.0) & ta (1.0) & ! (1.0) & sắc (1.0) & ? (1.0) & cảm (1.0) & đi (1.0) & tao (1.0) & đi (1.0) \\
vũ (.67) & hôm (1.0) & 1 (.5) & khẩu (1.0) & sợ (.67) & lắm (1.0) & mấy (.67) & lồn (.69) & mấy (.71) \\
bếp (.67) & đời (.6) & thành (.5) & trang (1.0) & ta (.67) & sợ (1.0) & tao (.67) & mấy (.62) & xe (.43) \\
vô (.67) & sống (.6) & ơn (.38) & 1 (1.0) & công (.67) & hung (.67) & mày (.33) & ta (.54) & ? (.43) \\
sống (.67) & thương (.6) & yêu (.38) & tao (1.0) & cúp (.67) & trọng (.67) & yêu (.33) & đi (.54) & bọn (.43) \\
\hline
\caption{Top 5 words causing misclassification for each emotion category (\colorbox{greenhead}{positive}, \colorbox{yellowhead}{ambiguous}, \colorbox{redhead}{negative}).}
\label{tab:misclass-words} \\
\end{longtable}
}

\clearpage
\onecolumn
\section{LLM prompt example}
\label{appendix_llm_prompt}
\subsection{Prompt \#1 - Vietnamese version}
\paragraph*{Objective:}
\noindent Classify Vietnamese social media comments into one or more of the 27 emotion categories defined by GoEmotions, plus neutral.

- - -
\paragraph*{Label Categories:}\

\begin{itemize}[noitemsep, topsep=0pt]
\item \noindent \textbf{amusement} (Giải trí) \includesvg[scale=0.015]{emojis/face-with-tears-of-joy} : Cảm giác buồn cười trước nội dung hài hước, thú vị. 
\item \textbf{excitement} (Hào hứng) \includesvg[scale=0.015]{emojis/star-struck} : Cảm giác vui mừng, phấn khích trước sự kiện, thông tin tích cực. 
\item \textbf{joy} (Niềm vui) \includesvg[scale=0.015]{emojis/grinning-face} : Cảm giác hạnh phúc, vui vẻ, thoải mái trước tình huống tích cực. 
\item \textbf{love} (Tình yêu) \includesvg[scale=0.015]{emojis/red-heart} : Bày tỏ yêu thương, gắn bó với người, vật, hoặc ý tưởng. 
\item \textbf{desire} (Mong muốn) \includesvg[scale=0.015]{emojis/smiling-face-with-heart-eyes} : Cảm giác mạnh mẽ về mong muốn có được ai đó hoặc điều gì đó. 
\item \textbf{optimism} (Lạc quan) \includesvg[scale=0.020]{emojis/crossed-fingers} : Hy vọng, tin tưởng vào kết quả tốt đẹp cho bản thân hoặc đối tượng khác. 
\item \textbf{caring} (Quan tâm) \includesvg[scale=0.115]{emojis/smiling-face-with-open-hands} : Thể hiện tình cảm, sự chăm sóc, hoặc hỗ trợ người khác. 
\item \textbf{pride} (Tự hào) \includesvg[scale=0.020]{emojis/relieved-face} : Hài lòng, kiêu hãnh về bản thân hoặc người khác. 
\item \textbf{admiration} (Ngưỡng mộ) \includesvg[scale=0.015]{emojis/clapping-hands} : Kính trọng, khâm phục ai đó vì phẩm chất hoặc thành tựu. 
\item \textbf{gratitude} (Biết ơn) \includesvg[scale=0.015]{emojis/folded-hands} : Cảm kích, trân trọng trước sự giúp đỡ, lời nói hoặc điều tích cực.
\item \textbf{relief} (Nhẹ nhõm) \includesvg[scale=0.015]{emojis/grinning-face-with-sweat} : Thở phào, giảm căng thẳng sau khi vượt qua vấn đề cho bản thân hoặc đối tượng khác.
\item \textbf{approval} (Chấp thuận) \includesvg[scale=0.020]{emojis/thumbs-up} : Đồng tình hoặc ủng hộ một ý kiến, hành động.
\item \textbf{realization} (Nhận ra) \includesvg[scale=0.015]{emojis/light-bulb} : Hiểu hoặc nhận thức, phát giác một điều gì đó.
\item \textbf{surprise} (Ngạc nhiên) \includesvg[scale=0.015]{emojis/astonished-face} : Bất ngờ trước thông tin hoặc sự kiện không lường trước.
\item \textbf{curiosity} (Tò mò) \includesvg[scale=0.018]{emojis/thinking-face} : Mong muốn tìm hiểu thêm, đặt câu hỏi, thắc mắc.
\item \textbf{confusion} (Bối rối) \includesvg[scale=0.015]{emojis/confused-face} : Lúng túng, không rõ ràng hoặc mâu thuẫn về thông tin.
\item \textbf{fear} (Sợ hãi) \includesvg[scale=0.020]{emojis/fearful-face} : Lo lắng, bất an trước điều nguy hiểm hoặc đáng sợ.
\item \textbf{nervousness} (Lo lắng) \includesvg[scale=0.018]{emojis/grimacing-face} : Căng thẳng, bồn chồn trước tình huống, sự kiện.
\item \textbf{remorse} (Hối hận) \includesvg[scale=0.018]{emojis/pensive-face} : Hối tiếc, tự trách về một quyết định hoặc hành động sai lầm.
\item \textbf{embarrassment} (Xấu hổ) \includesvg[scale=0.015]{emojis/flushed-face} : Ngượng ngùng, lúng túng trước tình huống khó xử, không phù hợp ở cá nhân hoặc đối tượng khác.
\item \textbf{disappointment} (Thất vọng) \includesvg[scale=0.015]{emojis/disappointed-face} : Hụt hẫng vì kỳ vọng không được đáp ứng.
\item \textbf{sadness} (Buồn bã) \includesvg[scale=0.015]{emojis/disappointed-face} : Thất vọng hoặc không vui.
\item \textbf{grief} (Đau buồn) \includesvg[scale=0.015]{emojis/crying-face} : Đau đớn hoặc buồn sâu sắc khi mất mát hoặc thất bại lớn.
\item \textbf{disgust} (Ghê tởm) \includesvg[scale=0.020]{emojis/face-vomiting} : Khó chịu, chê bai, chỉ trích, khinh bỉ với điều gì gây phản cảm.
\item \textbf{anger} (Tức giận) \includesvg[scale=0.415]{emojis/enraged-face} : Giận dữ, bức xúc vì bất công hoặc điều không vừa ý.
\item \textbf{annoyance} (Khó chịu) \includesvg[scale=0.015]{emojis/unamused-face} : Phiền toái trước điều gì đó không hài lòng.
\item \textbf{disapproval} (Phản đối) \includesvg[scale=0.018]{emojis/thumbs-down} : Không đồng ý, thể hiện sự không hài lòng.
\item \textbf{neutral} (Trung tính) : bình luận không thể hiện bất kỳ cảm xúc hay thái độ độ nào cụ thể. Nếu câu được gán neutral thì sẽ không gán các nhãn khác.
\end{itemize}

- - -
\paragraph*{Annotation Guidelines}\
\begin{itemize}[noitemsep, topsep=0pt]
\item Only use labels in label categories.
\item Select multiple labels if possible relevant.
\item No explanation
\end{itemize}

- - -
\paragraph*{Output example}\

\textbf{Label:} ["optimism", "confusion", "nervousness"]

- - -\\
Annotate the following text: \\
\texttt{<text>}

\subsection{Prompt \#2 - English version}

\paragraph*{Objective:}
\noindent Annotate Vietnamese social media comments with one or more of the 27 emotion categories from GoEmotions, plus neutral.

- - -
\paragraph*{Emotion Categories:}\

\begin{itemize}[noitemsep, topsep=0pt]
	\item \noindent \textbf{amusement} \includesvg[scale=0.015]{emojis/face-with-tears-of-joy} : Feeling entertained or amused by humorous or interesting content.
	\item \textbf{excitement} \includesvg[scale=0.015]{emojis/star-struck} : Feeling thrilled, enthusiastic, or highly interested in a positive event or information.
	\item \textbf{joy} \includesvg[scale=0.015]{emojis/grinning-face} : Feeling happy, cheerful, or satisfied in a positive situation.
	\item \textbf{love} \includesvg[scale=0.015]{emojis/red-heart} : Expressing affection, attachment, or strong emotional connection toward someone or something. 
	\item \textbf{desire} \includesvg[scale=0.015]{emojis/smiling-face-with-heart-eyes} : Strongly wanting someone or something.
	\item \textbf{optimism} \includesvg[scale=0.020]{emojis/crossed-fingers} : Hoping for a positive outcome for oneself or others.
	\item \textbf{caring} \includesvg[scale=0.115]{emojis/smiling-face-with-open-hands} : Expressing concern, care, or support for others.
	\item \textbf{pride} \includesvg[scale=0.020]{emojis/relieved-face} : Feeling satisfied or proud of oneself or others.
	\item \textbf{admiration} \includesvg[scale=0.015]{emojis/clapping-hands} : Showing respect or appreciation for someone’s qualities or achievements.
	\item \textbf{gratitude} \includesvg[scale=0.015]{emojis/folded-hands} : Expressing thankfulness or appreciation for help, kindness, or positive events.
	\item \textbf{relief} \includesvg[scale=0.015]{emojis/grinning-face-with-sweat} : Feeling reassured or at ease after overcoming a concern or problem.
	\item \textbf{approval} \includesvg[scale=0.020]{emojis/thumbs-up} : Agreeing with or supporting an idea, opinion, or action.
	\item \textbf{realization} \includesvg[scale=0.015]{emojis/light-bulb} : Suddenly understanding or discovering something.
	\item \textbf{surprise} \includesvg[scale=0.015]{emojis/astonished-face} : Expressing astonishment or unexpectedness at an event or information.
	\item \textbf{curiosity} \includesvg[scale=0.018]{emojis/thinking-face} : Showing interest in learning more or asking questions.
	\item \textbf{confusion} \includesvg[scale=0.015]{emojis/confused-face} : Feeling uncertain, unclear, or conflicted about information.
	\item \textbf{fear} \includesvg[scale=0.020]{emojis/fearful-face} : Feeling anxious or threatened by something scary or dangerous.
	\item \textbf{nervousness} \includesvg[scale=0.018]{emojis/grimacing-face} : Feeling tense, uneasy, or worried about a situation or event.
	\item \textbf{remorse} \includesvg[scale=0.018]{emojis/pensive-face} : Feeling regretful or guilty about a past decision or action.
	\item \textbf{embarrassment} \includesvg[scale=0.015]{emojis/flushed-face} : Feeling awkward or self-conscious in an uncomfortable situation.
	\item \textbf{disappointment} \includesvg[scale=0.015]{emojis/disappointed-face} : Feeling let down when expectations are not met.
	\item \textbf{sadness} \includesvg[scale=0.015]{emojis/disappointed-face} : Feeling sorrowful or unhappy.
	\item \textbf{grief} \includesvg[scale=0.015]{emojis/crying-face} : Experiencing deep sorrow due to a major loss or failure.
	\item \textbf{disgust} \includesvg[scale=0.020]{emojis/face-vomiting} : Expressing strong dislike, rejection, or criticism of something unpleasant.
	\item \textbf{anger} \includesvg[scale=0.415]{emojis/enraged-face} : Feeling upset or outraged about injustice or dissatisfaction.
	\item \textbf{annoyance} \includesvg[scale=0.015]{emojis/unamused-face} : Expressing irritation or mild displeasure.
	\item \textbf{disapproval} \includesvg[scale=0.018]{emojis/thumbs-down} : Showing disagreement or discontent.
	\item \textbf{neutral} : No specific emotion is conveyed. If "neutral" is assigned, no other labels should be selected.
\end{itemize}

- - -
\paragraph*{Annotation Guidelines}\
\begin{itemize}[noitemsep, topsep=0pt]
	\item Only use labels from the defined emotion categories.
	\item Assign multiple labels if applicable.
	\item No explanations required
\end{itemize}

- - -
\paragraph*{Output example}\

\textbf{Label:} ["nervousness", "embarrassment", "disappointment"]

- - -\\
Annotate the following text: \\
\texttt{<text>}

\section{Annotation Guidelines}
\label{appendix_annotation_guideline}
\subsection{Overview}
\textit{Objective:} annotate 27 emotion labels + \texttt{neutral}.\par
\textit{Problem type:} multi-class, multi-label.


\noindent \textbf{High-correlation label combination:}

(excitement, joy), (fear, nervousness), (disappointment, sadness, grief), (anger, annoyance).\\

\vspace{6pt}

\subsection{Class definition}
\paragraph{POSITIVE emotions}\

\noindent \textbf{Sorting of emotion categories}\\  
The emotion categories below are listed in ascending order according to each of the following criteria:

\begin{itemize}[noitemsep, topsep=0pt]
	\item \textit{Emotion intensity:} 
	approval; caring; optimism; relief; amusement; gratitude; admiration; pride; desire; excitement; joy; love.
	\item \textit{Context dependence:}
	caring; gratitude; pride; admiration; love; joy; relief; optimism; excitement; amusement; desire; approval.
\end{itemize}

\noindent \textbf{Definition}
\begin{itemize}[noitemsep, topsep=0pt]
	\item \textbf{amusement (giải trí) \includesvg[scale=0.015]{emojis/face-with-tears-of-joy}:}\\
	Cảm giác vui vẻ, hài hước khi thấy điều gì đó buồn cười hoặc thú vị.\\
	\textit{A sense of fun or humor when encountering something funny or entertaining.}
	\begin{itemize}[noitemsep, topsep=0pt]
		\item \textbf{Ví dụ:} "sau em này mà làm cô giáo ngồi kể chuyện hs lại há mồm ra nghe \includesvg[scale=0.015]{emojis/beaming-face-with-smiling-eyes}", "bài viết hay quá !"
		\item \textit{Example: “If she becomes a teacher one day, her students will be sitting there with their mouths wide open listening to her stories \includesvg[scale=0.015]{emojis/beaming-face-with-smiling-eyes}”, “Such a great article!”.}
	\end{itemize}

	\item \textbf{excitement (hào hứng) \includesvg[scale=0.015]{emojis/star-struck}:}\\
	Cảm giác phấn khởi hoặc hào hứng về một sự kiện, điều gì đó đang hoặc sẽ xảy ra.\\
	\textit{A feeling of enthusiasm or anticipation for something happening or about to happen.}
	\begin{itemize}
		\item \textbf{Ví dụ:} "dự định tao là 28 nhưng giờ mà có bồ là bắt cưới luôn chớ sợ ế haha"
		\item \textit{Example: “My plan was to get married at 28, but if I get a partner now, I’ll get married right away — not afraid of ending up single haha.”}
	\end{itemize}
	
	\item \textbf{joy (niềm vui) \includesvg[scale=0.015]{emojis/grinning-face}:}\\
	Cảm giác hạnh phúc, vui vẻ hoặc mãn nguyện.\\
	\textit{A feeling of happiness, delight, or contentment.}
	\begin{itemize}[noitemsep, topsep=0pt]
		\item \textbf{Ví dụ:} "mỗi ngày mình ôm nhau 1 cái như này hén , hí hí"
		\item \textit{Example: “Let’s hug like this every day, hehe.”}
	\end{itemize}

	\item \textbf{love (tình yêu) \includesvg[scale=0.015]{emojis/red-heart}:} \\
	Cảm giác yêu thương, chăm sóc và sẵn lòng hỗ trợ người khác.\\
	\textit{A deep sense of attachment, affection, and care toward someone or something.}
	\begin{itemize}
		\item \textbf{Ví dụ:} "mỗi lần có video của con là cứ coi đi coi lại hoài . cưng con quá ."
		\item \textit{Example: “Every time there’s a video of you, I just keep watching it over and over. Love you so much.”}
	\end{itemize}

	\item \textbf{desire (mong muốn) \includesvg[scale=0.015]{emojis/smiling-face-with-heart-eyes}:}\\
	Cảm giác mạnh mẽ về mong muốn có được ai đó hoặc điều gì đó.\\
	\textit{A strong feeling of wanting someone or something.}
	\begin{itemize}[noitemsep, topsep=0pt]
		\item\textbf{Ví dụ:} "uớc gì sau này về già vẫn có thể như cụ này :))"
		\item \textit{Example: “I wish I could be like this old man when I’m old :))"}
	\end{itemize}

	\item \textbf{optimism (lạc quan) \includesvg[scale=0.020]{emojis/crossed-fingers}:}\\
	Biểu hiện của hy vọng, tin tưởng vào kết quả tốt đẹp hoặc nhìn nhận tích cực dù trong khó khăn. Tính chất: cường độ thấp; phụ thuộc bối cảnh mức trung bình; hướng đến tình huống.\\
	\textit{“The expression of a feeling of hope, trust in a positive outcome, or viewing a situation in a positive light, even when facing difficulties. Comments with an optimistic tone can encourage, motivate, or reflect belief in the possibility of positive change, whether for oneself, others, or a specific situation.”}
	\begin{itemize}
		\item \textbf{Ví dụ:} "thích chị này nhất phim í , bé bé xinh xinh lại diễn rất đạt :3 . hy vọng mai sau sẽ có bước tiến lớn và thành công hơn nữa <3"; "cố gắng lên mọi lỗ lực và công lao của em sẽ được đền đáp một cách thật là xứng đáng , cố gắng lên em nhé \includesvg[scale=0.015]{emojis/thumbs-up}"
		\item \textit{Example: "I like this actress the most in the movie; she is cute, petite, and acts very well :3. I hope she will make big progress and be even more successful in the future <3."}; \textit{"Keep trying; all your effort and hard work will be rewarded truly and deservedly. Keep going \includesvg[scale=0.015]{emojis/thumbs-up}"}
	\end{itemize}

	\item \textbf{caring (quan tâm) \includesvg[scale=0.10]{emojis/smiling-face-with-open-hands}:}\\
	Cảm giác yêu thương, chăm sóc và sẵn lòng hỗ trợ người khác.\\
	\textit{A feeling of affection, care, and willingness to help others.}
	\begin{itemize}
		\item \textbf{Ví dụ:} "thật tốt vì chú chó ấy đã tìm được người yêu thương nó , kiên trì để nó tin tưởng thật sự :">"
		\item \textit{Example: “It’s really great that the dog has found someone who loves it and is patient enough for it to truly trust them :”>”}
	\end{itemize}

	\item \textbf{pride (tự hào) \includesvg[scale=0.020]{emojis/relieved-face}:}\\
	Cảm giác tự tin, hài lòng hoặc kiêu hãnh về thành quả, phẩm chất hoặc sự đóng góp của cá nhân hoặc một nhóm. Cảm giác này có thể xuất hiện ngay cả trong lời khen ngợi, sự công nhận ngầm, hoặc sự hài lòng khi chứng kiến thành công hoặc giá trị đạt được, ngay cả khi không nêu rõ cụ thể.\\
	\textit{“A feeling of confidence, satisfaction, or pride in one’s achievements, qualities, or contributions, whether individually or as part of a group. This feeling may also appear in compliments, subtle recognition, or the sense of fulfillment when witnessing success or value being achieved, even when not stated explicitly.”}
	\begin{itemize}
		\item \textbf{Ví dụ:} "bản thân quá hạnh phúc khi trong cs này rất . rất nhiều người mơ ước được như mình hiện tại . \includesvg[scale=0.020]{emojis/smiling-face-with-smiling-eyes}\includesvg[scale=0.020]{emojis/smiling-face-with-smiling-eyes}\includesvg[scale=0.020]{emojis/smiling-face-with-smiling-eyes}"
		\item \textit{Example: “I’m so happy because in this life, there are very… very many people who dream of being where I am right now. ”}
	\end{itemize}

	\item \textbf{admiration (ngưỡng mộ) \includesvg[scale=0.015]{emojis/clapping-hands}:}\\
	Cảm giác kính trọng hoặc ngưỡng mộ khả năng hoặc phẩm chất của ai đó.\\
	\textit{A feeling of respect or admiration for someone's abilities or qualities.}
	\begin{itemize}
		\item \textbf{Ví dụ:} "thấy bạn kia cứu con chó không cần nhìn mặt cũng biết đẹp trai rồi"
		\item \textit{Example: “Seeing that guy save the dog—don’t need to see his face to know he’s handsome.”}
	\end{itemize}

	\item \textbf{gratitude (biết ơn) \includesvg[scale=0.015]{emojis/folded-hands}:}\\
	Đây là trạng thái cảm xúc xuất hiện khi một người nhận ra và đánh giá cao những điều tốt đẹp mà họ có được — dù những điều đó đến từ người khác, từ hoàn cảnh, từ môi trường, hay từ những trải nghiệm tích cực nói chung. Cảm giác biết ơn thường đi kèm sự ấm áp và sự ghi nhận rằng điều tích cực ấy có ý nghĩa đối với bản thân.\\
	\textit{This emotional state arises when someone recognizes and values the positive aspects they have received or experienced — whether they come from other people, circumstances, the environment, or any beneficial event. Gratitude is often accompanied by a sense of warmth and an acknowledgment of the meaningful impact of that positive experience.}
	\begin{itemize}
		\item \textbf{Ví dụ:} "cảm ơn nói chuyện nhật ! 1 nền khoa học , giáo dục , trí tuệ văn minh !! \includesvg[scale=0.020]{emojis/face-blowing-a-kiss}"
		\item \textit{Example: “Thank you, Japan, for the conversation! A nation of science, education, and civilized intelligence!! \includesvg[scale=0.020]{emojis/face-blowing-a-kiss}”}
	\end{itemize}

	\item \textbf{relief (nhẹ nhõm) \includesvg[scale=0.015]{emojis/grinning-face-with-sweat}:}\\
	Cảm giác thở phào, bớt căng thẳng hoặc lo lắng của bản thân hoặc đối tượng khác. Cảm giác biểu hiện qua các câu diễn tả sự an tâm, yên tâm, hoặc vui mừng ("thật tốt", "mừng",...) khi vượt qua một vấn đề nào đó.\\
	\textit{A feeling of relief or reduced tension or anxiety, either for oneself or for someone else. This emotion is expressed through statements that convey reassurance, ease, or gladness (such as ‘that’s good’ or ‘glad’) after overcoming a certain problem.}
	\begin{itemize}
		\item \textbf{Ví dụ:} "hên là chạy ra kịp", "tao quen rồi \includesvg[scale=0.020]{emojis/relieved-face}"
		\item \textit{Example: “Lucky I ran out in time”, “I’m used to it already \includesvg[scale=0.020]{emojis/relieved-face}”}
	\end{itemize}

	\item \textbf{approval (chấp thuận) \includesvg[scale=0.015]{emojis/thumbs-up}:}\\
	Cảm giác đồng tình hoặc ủng hộ một hành động, ý kiến, hoặc tình huống. Cảm giác này có thể thể hiện qua các bình luận tích cực, ngắn gọn hoặc chỉ ra sự đồng ý một cách gián tiếp, dù không nhất thiết nêu rõ lý do, và có thể kèm các từ ngữ hoặc biểu tượng thể hiện sự hài lòng.\\
	\textit{A feeling of agreement or support toward an action, opinion, or situation. This emotion may be expressed through positive or concise comments that indicate indirect approval, without necessarily stating the reason, and may include words or symbols that show satisfaction.}
	\begin{itemize}
		\item \textbf{Ví dụ:} "tôi ở phú yên và anh này nói rất chuẩn < 3"
		\item \textit{Example: “I’m in Phú Yên, and this guy speaks so accurately <3”}
	\end{itemize}

\end{itemize}

\paragraph{AMBIGUOUS emotions}\

\noindent \textbf{Sorting of emotion categories}\\  
The emotion categories below are listed in ascending order according to each of the following criteria:

\begin{itemize}[noitemsep, topsep=0pt]
	\item \textit{Emotion intensity:} curiosity; confusion; surprise; realization.
	\item \textit{Context dependence:} curiosity; confusion; surprise; realization.
\end{itemize}

\noindent \textbf{Definition}

\begin{itemize}[noitemsep, topsep=0pt]

	\item \textbf{realization (nhận ra) \includesvg[scale=0.015]{emojis/light-bulb}:}\\
	Cảm giác hiểu, nhận thức hoặc khám phá ra điều gì đó. Cảm giác này thường xuất hiện trong các bình luận chứa những cụm từ như "giờ mới hiểu," "thì ra là vậy," "đúng là như thế," hoặc bất kỳ cách diễn đạt nào thể hiện sự nhận thức mới hoặc sự bất ngờ về một sự thật mà trước đây không nhận ra. \\
	\textit{A feeling of understanding, realization, or discovery. This emotion often appears in comments containing phrases like ‘just realized,’ ‘so that’s how it is,’ ‘that’s true,’ or any expression that reflects a new awareness or surprise about a fact previously unnoticed.}
	\begin{itemize}
		\item \textbf{Ví dụ:} "tiêu đề bộ phim cứ ngỡ là thể loại hài , nhưng nếu xem hết thì các bạn sẽ rút ra rất nhiều điều và cái hay trong phim này"
		\item \textit{Example: “The movie title makes you think it’s a comedy, but if you watch it all the way through, you’ll take away a lot of insights and see what’s great about this film.”}
	\end{itemize}

	\item \textbf{surprise (ngạc nhiên) \includesvg[scale=0.015]{emojis/astonished-face}:}\\
	Cảm giác bất ngờ hoặc kinh ngạc trước một sự kiện hoặc thông tin không lường trước.\\
	\textit{A feeling of surprise or amazement at an unexpected event or piece of information.}
	
	\begin{itemize}
		\item \textbf{Ví dụ:} "thật hay đùa vậy, không thể tin được"
		\item \textit{Example: “Are you serious or joking? I can’t believe it.”}
	\end{itemize}

	\item \textbf{curiosity (tò mò) \includesvg[scale=0.015]{emojis/thinking-face}}\\
	Cảm giác muốn tìm hiểu hoặc biết một điều gì đó, thể hiện sự quan tâm đến thông tin mới hoặc chưa biết. Cảm xúc này thường xuất hiện trong các bình luận thể hiện mong muốn tìm hiểu thêm hoặc có nhu cầu giải đáp thắc mắc.\\
	\textit{A feeling of curiosity or a desire to learn something, reflecting interest in new or unknown information. This emotion often appears in comments expressing a wish to explore further or a need for answers.}
	\begin{itemize}
		\item \textbf{Ví dụ:} thế ông bác sĩ đó bây giờ còn sống không \includesvg[scale=0.015]{emojis/slightly-smiling-face}"
		\item \textit{Example: “Is that doctor still alive now? \includesvg[scale=0.015]{emojis/slightly-smiling-face}”}
	\end{itemize}

	\item \textbf{confusion (bối rối) \includesvg[scale=0.015]{emojis/confused-face}:}\\
	Cảm giác lúng túng hoặc không chắc chắn về điều gì đó, thường do thiếu thông tin hoặc thông tin mâu thuẫn.\\
	\textit{A feeling of confusion or uncertainty about something, often due to a lack of information or conflicting information.}
	\begin{itemize}
		\item \textbf{Ví dụ:} "đây là đâu ? cái này là như nào thế ??"
		\item \textit{Example: “Where is this? What is this supposed to be??”}
	\end{itemize}

\end{itemize}

\paragraph{NEGATIVE emotions}\

\noindent \textbf{Sorting of emotion categories}\\  
The emotion categories below are listed in ascending order according to each of the following criteria:

\begin{itemize}[noitemsep, topsep=0pt]
	\item \textit{Emotion intensity:} disapproval; annoyance; embarassment; nervousness; disgust; fear; disappointment; sadness; remorse; grief; anger.
	\item \textit{Context dependence:} sadness; remorse; grief; anger; disappointment; fear; disgust; disapproval; annoyance; nervousness; embarrassment.
\end{itemize}

\noindent \textbf{Definition}

\begin{itemize}[noitemsep, topsep=0pt]

	\item \textbf{fear (sợ hãi) \includesvg[scale=0.018]{emojis/fearful-face}:}\\
	Cảm giác lo lắng hoặc hoảng loạn khi đối mặt với điều gì đó nguy hiểm, đe dọa hoặc không rõ ràng. Cảm giác này thường gắn liền với sự phản ứng mạnh mẽ đối với mối nguy hiểm trực tiếp và có thể dẫn đến hành động né tránh hoặc tìm kiếm sự an toàn.\\
	\textit{A feeling of worry or panic when facing something dangerous, threatening, or uncertain. This emotion is often associated with a strong reaction to immediate danger and may lead to avoidance behavior or seeking safety.}
	\begin{itemize}
		\item \textbf{Ví dụ:} "từ đầu năm giờ chắc chụp 10 lần rồi huhu sợ quá \includesvg[scale=0.015]{emojis/loudly-crying-face}\includesvg[scale=0.015]{emojis/loudly-crying-face}\includesvg[scale=0.015]{emojis/loudly-crying-face}"
		\item \textit{Example: “Since the start of the year I’ve taken X-rays like 10 times, so scared \includesvg[scale=0.015]{emojis/loudly-crying-face}\includesvg[scale=0.015]{emojis/loudly-crying-face}\includesvg[scale=0.015]{emojis/loudly-crying-face}”}
	\end{itemize}

	\item \textbf{nervousness (lo lắng) \includesvg[scale=0.018]{emojis/grimacing-face}:}\\
	Cảm giác bất an, bồn chồn trước một tình huống căng thẳng hoặc không chắc chắn, thường liên quan đến lo lắng về phản ứng của người khác hoặc kết quả của một sự kiện. Không nhất thiết phải gắn liền với mối nguy hiểm cụ thể, mà có thể là sự lo lắng về những điều nhỏ nhặt trong cuộc sống hàng ngày.\\
	\textit{A feeling of unease or restlessness in a stressful or uncertain situation, often related to concern about others’ reactions or the outcome of an event. It does not necessarily involve a specific danger, and can be worry about minor things in everyday life.}
	\begin{itemize}
		\item \textbf{Ví dụ:} "tao học nhân sự đây còn đang lo thất nghiệp đây này"
		\item \textit{Example: “I study HR and I’m still worried about being unemployed.”}
	\end{itemize}

	\item \textbf{remorse (hối hận) \includesvg[scale=0.015]{emojis/pensive-face}:}\\
	Cảm giác hối tiếc hoặc tự trách về một hành động, quyết định hoặc lời nói, có thể không cần nhấn mạnh vào chi tiết cụ thể nhưng biểu lộ sự nhận thức về sai lầm. Hối hận thường đi kèm với cảm giác áy náy hoặc băn khoăn khi nhìn lại tình huống.\\
	\textit{A feeling of regret or self-reproach about an action, decision, or statement, which may not emphasize specific details but expresses awareness of a mistake. Regret is often accompanied by feelings of guilt or unease when reflecting on the situation.}
	\begin{itemize}
		\item \textbf{Ví dụ:} "bà tôi giờ mất mấy năm rồi giờ lớn lên rồi mà chả tặng được bà món quà nào cả \includesvg[scale=0.015]{emojis/loudly-crying-face}\includesvg[scale=0.015]{emojis/loudly-crying-face}\includesvg[scale=0.015]{emojis/loudly-crying-face}"
		\item \textit{Example: “My grandmother passed away several years ago, and now that I’ve grown up, I still haven’t been able to give her any gift \includesvg[scale=0.015]{emojis/loudly-crying-face}\includesvg[scale=0.015]{emojis/loudly-crying-face}\includesvg[scale=0.015]{emojis/loudly-crying-face}”}
	\end{itemize}

	\item \textbf{embarrassment (xấu hổ) \includesvg[scale=0.015]{emojis/flushed-face}:}\\
	Cảm giác ngượng ngùng, lúng túng hoặc khó xử, có thể xuất phát từ tình huống cá nhân hoặc do chứng kiến người khác làm điều gì không phù hợp hoặc khó coi. Cảm xúc này bao gồm cả khi cá nhân cảm thấy xấu hổ dùm cho đối phương, như khi chứng kiến hành động gây khó xử của người khác hoặc khi một ai đó bị chỉ trích, thu hút sự chú ý không mong muốn.\\
	\textit{A feeling of embarrassment, awkwardness, or discomfort, which may arise from a personal situation or from witnessing someone else doing something inappropriate or awkward. This emotion also includes feeling embarrassed on behalf of someone else, such as when observing another person’s awkward actions or when someone is criticized or draws unwanted attention.}
	\begin{itemize}
		\item \textbf{Ví dụ:} "chỉ mới hỏi về người khác giới cái là bị gán là thích bạn đó luôn mí sợ :("
		\item \textit{Example: “Just asking about someone of the opposite gender and people immediately assume I like them, so scary :(”}
	\end{itemize}

	\item \textbf{disappointment (thất vọng) \includesvg[scale=0.015]{emojis/disappointed-face}:}\\
	Cảm giác hụt hẫng khi kỳ vọng không được đáp ứng.\\
	\textit{A feeling of letdown when expectations are not met.}
	\begin{itemize}
		\item \textbf{Ví dụ:} "đến năm cuối đại học rồi lúc nhớ lại đéo có gì :))"
		\item \textit{Example: “In my final year of university and remembering I achieved nothing :))”}
	\end{itemize}

	\item \textbf{sadness (buồn bã) \includesvg[scale=0.015]{emojis/disappointed-face}:}\\
	Cảm giác đau lòng hoặc không vui.\\
	\textit{A feeling of sorrow or unhappiness.}
	\begin{itemize}
		\item \textbf{Ví dụ:} "ngày buồn nhất trời đổ cơn mưa"
		\item \textit{Example: “On the saddest day, it rained.”}
	\end{itemize}

	\item \textbf{grief (đau buồn) \includesvg[scale=0.015]{emojis/crying-face}:}\\
	Cảm giác đau đớn sâu sắc khi mất mát.\\
	\textit{A deep sense of sorrow over loss.}
	\begin{itemize}
		\item \textbf{Ví dụ:} "đã khóc . cha mẹ là người luôn yêu thương minh nhất", "nhìn thôi đã đau lòng";
		\item \textit{Example: “I cried. Parents are the ones who love us the most,” “Just looking at it already hurts.”}
	\end{itemize}

	\item \textbf{disgust (ghê tởm) \includesvg[scale=0.025]{emojis/face-vomiting}:}\\
	Cảm giác khó chịu, không thích hoặc khinh bỉ điều phản cảm.\\
	\textit{A feeling of revulsion or strong dislike toward something unpleasant.}
	\begin{itemize}
		\item \textbf{Ví dụ:} "cho đáng đời con quỷ . về nhà lôi con nhà mày ra mà đánh \includesvg[scale=0.515]{emojis/enraged-face}"
		\item \textit{Example: “Serves that demon right. Go home and drag your kid out to beat them \includesvg[scale=0.515]{emojis/enraged-face}”}
	\end{itemize}

	\item \textbf{anger (tức giận) \includesvg[scale=0.515]{emojis/enraged-face}:}\\
	Cảm giác giận dữ hoặc bực bội khi gặp điều sai trái.\\
	\textit{A feeling of anger or frustration at something unfair or wrong.}
	\begin{itemize}
		\item \textbf{Ví dụ:} "per mày cẩn thận tao giết"
		\item \textit{Example: “Per, watch out or I’ll kill you”}
	\end{itemize}

	\item \textbf{annoyance (khó chịu) \includesvg[scale=0.015]{emojis/unamused-face}:}\\
	Cảm giác phiền toái, khó chịu với ai đó hoặc điều gì đó.\\
	\textit{A feeling of irritation or bother toward someone or something.}
	\begin{itemize}
		\item \textbf{Ví dụ:} "nó bướng vãi, cứ làm theo ý mình"
		\item \textit{Example: “They’re so stubborn, always doing things their way.”}
	\end{itemize}

	\item \textbf{disapproval (Phản đối) \includesvg[scale=0.015]{emojis/thumbs-down}:}\\
	Cảm giác không đồng ý hoặc không hài lòng về hành động, ý kiến hoặc tình huống.\\
	\textit{A feeling of disagreement or dissatisfaction toward an action, opinion, or situation.}
	\begin{itemize}
		\item \textbf{Ví dụ:} "ổng quay thiệt mà sao chửi ổng thế . không quay sao máy bạn xem được đó :))"
		\item \textit{Example: “He really did film it, so why are you cursing him? If he didn’t film it, how would you be watching it? :))”}
	\end{itemize}

\end{itemize}

\paragraph{NEUTRAL}\

\begin{itemize}[noitemsep, topsep=0pt]
	\item \textbf{neutral (trung tính)}:\\ 
	Những bình luận không thể hiện bất kỳ cảm xúc hay thái độ độ nào cụ thể. Nếu câu được gán neutral thì sẽ không gán thêm các nhãn khác. Một số dấu hiệu để nhận biết neutral:\\
	\textit{Comments that do not express any specific emotion or attitude. If a sentence is labeled neutral, no additional emotion labels should be applied. Some indicators for identifying neutral comments include:}
	\begin{itemize}[noitemsep, topsep=0pt]
		\item Không có ngôn ngữ cảm xúc hoặc từ ngữ thể hiện thái độ. / \textit{No emotional language or words expressing attitude.}
		\item Không sử dụng các emoji hoặc dấu câu biểu hiện sự biểu cảm. / \textit{No use of emojis or punctuation that conveys expressiveness.}
		\item Nội dung mang tính chất thông tin, khách quan, hoặc chỉ đơn thuần mô tả mà không có yếu tố chủ quan hay cảm xúc. / \textit{Content that is informational, objective, or purely descriptive without subjective or emotional elements.}
	\end{itemize}

\end{itemize}

\subsection{Labeling Method}

\begin{itemize}
	\item Initially, each sample is assigned a preliminary set of labels by the LLM. These labels serve as an initial suggestion to support the annotation process.
    \item Then, each annotator independently reviews the label set proposed by the LLM: keeping labels that are correct, adding any missing labels, and removing labels they judge to be inappropriate. Annotators may assign multiple labels based on the label definitions and their perception of the sample.
	\item Finally, for each sample, using the labeling results from the three annotators (or more), each label is considered separately. If a label is selected by at least two annotators, that label will be chosen. For example:
\end{itemize}

\begin{center}
	\begin{tabular}{|m{2cm}|m{6cm}|}
		\hline
		\textbf{id} & \textbf{comment} \\
		\hline
		ee6lqiq & \texttt{<comment\_content>} \\
		\hline
	\end{tabular}
\end{center}

\noindent
\textbf{Ratings:}
\begin{itemize}
	\item rater 1: [14, 19, 25]
	\item rater 2: [3, 12, 19, 25]
	\item rater 3: [1, 3, 11, 15]
\end{itemize}

\noindent
\textbf{Final labels:} [3, 19, 25]

\end{document}